\documentclass[11pt]{article}

\usepackage[preprint]{acl}

\usepackage{times}
\usepackage{latexsym}

\usepackage[T1]{fontenc}

\usepackage[utf8]{inputenc}

\usepackage{microtype}

\usepackage{inconsolata}

\usepackage{graphicx}

\usepackage{xcolor}

\usepackage{amsmath}
\usepackage{hyperref}
\usepackage{url}
\usepackage{algorithm}
\usepackage{algorithmicx}
\usepackage{algpseudocode}
\usepackage{booktabs}
\usepackage{multirow}
\usepackage[table]{xcolor}
\usepackage{cleveref}
\usepackage{wrapfig}
\usepackage{float}
\usepackage{amssymb}
\usepackage[most]{tcolorbox}

\newtcolorbox{examplebox}[1]{
    colback=white,
    colframe=black!10,
    coltitle=black,
    fonttitle=\bfseries,
    title=#1,
    sharp corners,
    boxrule=0.6pt,
    width=\linewidth, 
    enhanced,
    unbreakable, 
    before skip=10pt,
    after skip=10pt,
    fontupper=\small
}

\title{\fontsize{18}{22}\selectfont \textbf{Beyond Hard Masks: Progressive Token Evolution for Diffusion Language Models}}

\author{
\begin{tabular}{l}
Linhao Zhong$^{1}$\thanks{Equal Contribution.} \quad Linyu Wu$^{2}$\footnotemark[1] \quad Bozhen Fang$^{1}$ \quad Tianjian Feng$^{1}$ \quad Chenchen Jing$^{1,3}$ \\ Wen Wang$^{1}$ \quad Jiaheng Zhang$^{2}$ \quad Hao Chen$^{1}$  \quad Chunhua Shen$^{1,3}$\thanks{Corresponding Author.} \\
\textnormal{$^{1}$Zhejiang University} \quad
\textnormal{$^{2}$National University of Singapore} \quad
\textnormal{$^{3}$Zhejiang University of Technology}
\end{tabular}
}

\begin{document}
\maketitle
\begin{abstract}

Diffusion Language Models (DLMs) offer a promising alternative for language modeling by enabling parallel decoding through iterative refinement.
However, most DLMs rely on hard binary masking and discrete token assignments, which hinder the revision of early decisions and underutilize intermediate probabilistic representations.
In this paper, we propose EvoToken-DLM, a novel diffusion-based language modeling approach that replaces hard binary masks with evolving soft token distributions.
EvoToken-DLM enables a progressive transition from masked states to discrete outputs, supporting revisable decoding. 
To effectively support this evolution, we introduce continuous trajectory supervision, which aligns training objectives with iterative probabilistic updates.
Extensive experiments across multiple benchmarks show that EvoToken-DLM consistently achieves superior performance, outperforming strong diffusion-based and masked DLM baselines.
Project webpage: \url{https://aim-uofa.github.io/EvoTokenDLM}.

\end{abstract}
\section{Introduction}

\begin{figure}[t]
    \centering
    \includegraphics[width=\linewidth]{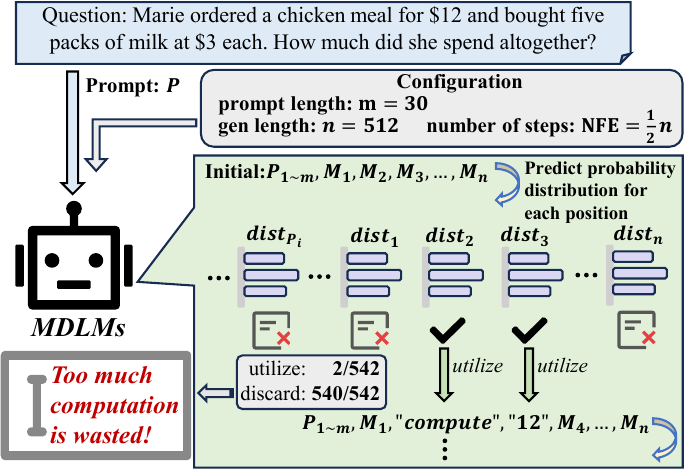}
    \vspace{-8mm}
    \caption{
    Inefficient utilization of predictions in masked diffusion language models, where distributions are computed for all positions but only a subset are used for decoding.
    $[M_1, M_2, \dots, M_n]$ denote the initial mask tokens following prompt $P$, and $dist_i$ represents the predicted probability distribution for the $i$-th token in the generation sequence.
    In this example, the total sequence of $542$ tokens consists of $30$ prompt tokens and $512$ generated tokens, while only two positions are updated per step.
    }
    \label{fig:waste-computation}
\end{figure}


Diffusion Language Models (DLMs)~\citep{llada, llada1.5, dream2025} frame language generation as an iterative refinement process, enabling parallel decoding in contrast to the strictly sequential nature of autoregressive models.
By replacing causal token-by-token generation with diffusion-based refinement~\citep{DDPM, IDDPM, DDIM}, DLMs offer an alternative generation paradigm that improves decoding parallelism. 

Most existing DLMs adopt a masked diffusion fashion, commonly referred to as masked diffusion language models (MDLMs) \citep{llada, llada1.5, dream2025}, in which generation is performed by maintaining a partially masked sequence and progressively replacing masked positions with discrete token assignments, enabling the simultaneous decoding of multiple tokens. 
To further improve the practicality of DLMs, recent work has introduced KV-caching mechanisms~\citep{wu2025fast, wu2025fast2, ma2025dkv, chen2025dparallel} that reuse hidden states across refinement steps to reduce redundant computation. In parallel, blockwise diffusion models~\citep{d2f, sdar, llada2} apply diffusion-based generation within local token blocks while preserving autoregressive dependencies across blocks, combining global causal coherence with local parallel efficiency.

\begin{figure*}[t]
    \centering
    \includegraphics[width=\linewidth]{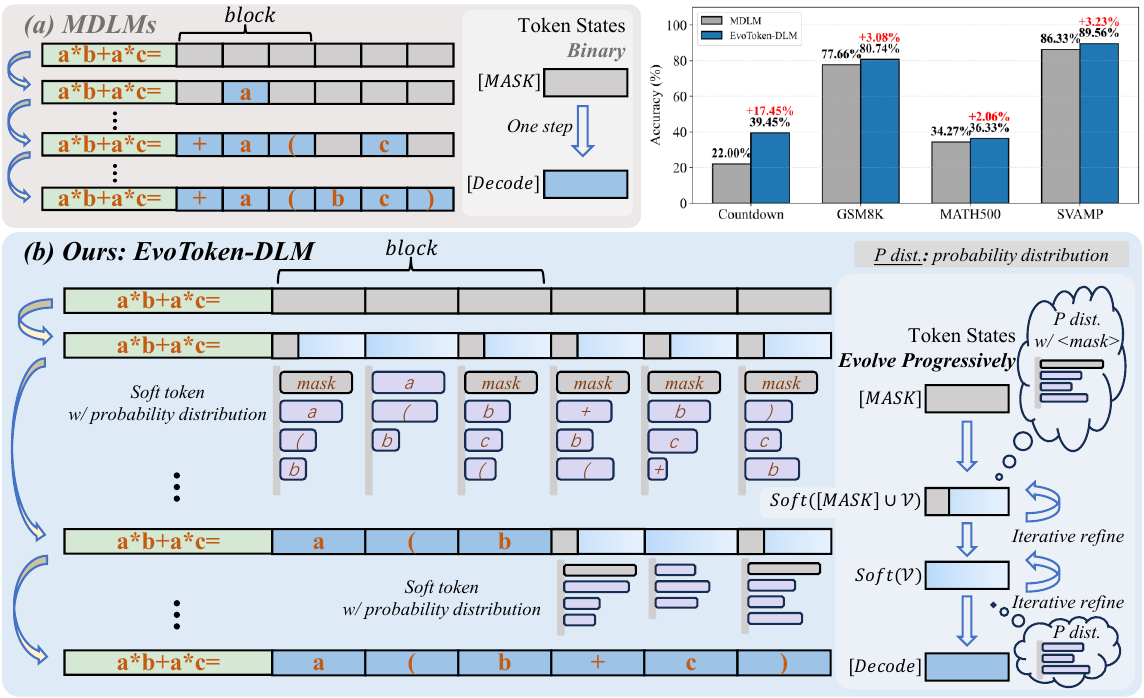}
    \vspace{-8mm}
    \caption{
    Comparison between MDLMs and EvoToken-DLM.
    (a) Standard MDLMs employ only two token states, alternating between \texttt{<mask>} and discrete decoded tokens, leading to abrupt mask-to-token transitions.
    (b) EvoToken-DLM introduces soft tokens represented by probability distributions and four token states, enabling tokens to evolve progressively through iterative refinement.
    The top-right panel illustrates a quantitative comparison between the two approaches under the same settings based on LLaDA-Instruct-8B.
    }
    \label{fig:teaser}
\end{figure*}

However, most MDLMs rely on hard binary masking with discrete token assignments. Once a token is decoded, it is treated as final and excluded from further refinement, resulting in an abrupt transition from uncertainty to determinism. This irreversibility limits the model's ability to revise early decisions and undermines the iterative refinement paradigm of diffusion-based language modeling.
In addition, as illustrated in Figure~\ref{fig:waste-computation}, although MDLMs compute token distributions for all positions at each refinement step, only a small subset of positions are updated, while the remaining probabilistic information is discarded.

In this work, we propose \textbf{EvoToken-DLM}, a diffusion-based language modeling approach that replaces hard binary masks with evolving soft token distributions. 
Instead of predicting discrete tokens in a single step, EvoToken-DLM represents each token as a probability distribution over the vocabulary and iteratively refines it throughout the diffusion process. 
As illustrated in Figure~\ref{fig:teaser}, token decoding becomes a progressive and continuous evolution:
\begin{equation}
\begin{aligned}
\texttt{[MASK]}
&\;\rightarrow\;
\mathrm{Soft}(\texttt{[MASK]} \cup \mathcal{V}) \\
&\;\rightarrow\;
\mathrm{Soft}(\mathcal{V})
\;\rightarrow\;
\texttt{[Decode]} .
\end{aligned}
\end{equation}
This evolution gradually transitions tokens from masked uncertainty to mask-aware soft token distributions, then to fully soft token distributions, and finally to discrete outputs. By allowing token representations to evolve across refinement steps before being finalized, EvoToken-DLM enables smooth and revisable decoding, mitigating premature decisions induced by hard masking.

To support this progressive refinement during training, we introduce \textbf{continuous trajectory supervision}, which aligns training objectives with iterative probabilistic updates along the diffusion trajectory. 
EvoToken-DLM requires no modification to the underlying model architecture and can be readily adapted from existing MDLMs. 
Moreover, it is fully compatible with KV-caching and naturally extends to blockwise diffusion settings, demonstrating broad applicability. Extensive experiments across multiple benchmarks show that EvoToken-DLM consistently outperforms strong MDLM baselines.




Our main contributions are:
\begin{itemize}
    \item We propose EvoToken-DLM, a diffusion-based language modeling approach that replaces hard binary masks with evolving soft token distributions, enabling a staged and revisable decoding process throughout diffusion.

    \item We introduce a continuous trajectory supervision-based training strategy that aligns model optimization with iterative probabilistic token refinement along the diffusion trajectory, effectively supporting progressive token evolution.

    \item We demonstrate that EvoToken-DLM integrates seamlessly with KV-caching and extends naturally to blockwise diffusion architectures. Extensive experiments across diverse model backbones, datasets, and inference configurations show consistent and robust improvements over strong MDLM baselines, highlighting EvoToken-DLM as a general and effective enhancement.
    
\end{itemize}

\section{Preliminaries on MDLMs}

Masked diffusion language models operate under a masked diffusion paradigm. The generation process consists of two main stages: a forward corruption process and a learned reverse denoising process.

\paragraph{Forward Process.} Given an original text sequence $X^0 = (x_1^0, \dots, x_N^0)$ of $N$ tokens, the forward process gradually corrupts it into a noisy sequence $X^t$ over a time schedule $t \in [0,T]$. This corruption is typically achieved by independently replacing each token with a special mask token with probability $\frac{t}{T}$:
\[
q(x_i^t \mid x_i^0) =
\begin{cases}
1-\frac{t}{T}, & \text{if } x_i^t = x_i^0, \\
\frac{t}{T}, & \text{if } x_i^t = \texttt{<mask>}
\end{cases}
\]
At $t=T$, the sequence $X^T$ becomes fully masked.

\paragraph{Reverse Process.} MDLMs learn a parameterized model $p_\theta(X^0 \mid X^t)$ to reverse the forward process. The model predicts all masked tokens simultaneously at each inference step, enabling high-speed, parallel generation from the fully masked sequence $X^T$ to the original text $X^0$. 

In practice, at each decoding step, the model selects a subset of masked tokens to finalize based on their predicted confidence rather than decoding all tokens at once. Masked tokens not selected in the current step remain in the mask state and will be decoded in subsequent steps. 
Furthermore, sequences are partitioned into discrete blocks that are processed in a sequential manner, where the model advances to the next block only upon the complete refinement of all masked tokens within the current one.
\section{From Discrete to Continuous: A Continuous Relaxation Perspective}

\paragraph{Continuous Relaxation.} 
Let $\mathcal{V} = \{1, \dots, V\}$ denote the vocabulary of size $V$. We define the discrete token space as the set of one-hot vectors $\mathcal{X} = \{\delta_1, \dots, \delta_V\} \subset \{0,1\}^V$. Associated with the vocabulary is an embedding matrix $\mathbf{U} \in \mathbb{R}^{V \times D}$. The embedding function maps a token index $i$ to a continuous vector $\mathbf{e} = \mathbf{U}_i$.
We denote the continuous embedding space as the convex hull of the token embeddings: $\mathcal{E} = \text{Conv}(\mathbf{U}) \subset \mathbb{R}^D$. A soft token is any vector $\tilde{\mathbf{e}} \in \mathcal{E}$ that can be expressed as $\tilde{\mathbf{e}} = \mathbf{U}^\top \mathbf{p}$, where $\mathbf{p} \in \Delta^{V-1}$ lies on the probability simplex.
This formulation relaxes the categorical selection into a continuous domain.

\paragraph{Iterative Refinement in Continuous Domain.} 
Unlike standard MDLMs which predict $p_\theta(X^0 \mid X^T)$ iteratively over the discrete vocabulary, our method models the reverse process as an iterative refinement loop in the continuous domain $\mathcal{E}$. 
Specifically, let $X^T$ denote the masked input sequence and $\mathbf{E}^T$ be its corresponding embeddings, where each element of $\mathbf{E}^T$ belongs to $\mathcal{E}$.
We introduce auxiliary token states $\mathbf{Z}^T$ to enable continuous token evolution. The refinement process is governed by a transition function $\Phi$, which recursively updates both the continuous embeddings and the token states: $(\mathbf{E}^{t-1}, \mathbf{Z}^{t-1}) = \Phi(\mathbf{E}^{\geq t}, \mathbf{Z}^{\geq t})$. Through successive applications of $\Phi$, the model progressively purifies the noisy input until it reaches the terminal $\mathbf{E}^0, \mathbf{Z}^0$. 
Finally, $\mathbf{E}^0$ is mapped back to the discrete domain to produce the output sequence $X^0$.

\section{EvoToken-DLM}



\subsection{Progressive Inference with EvoToken-DLM}
\label{sec:progressive_inference}


We formally define the progressive inference procedure of EvoToken-DLM as follows. Given a prompt $P$, the objective is to generate a response of length $N$. The output is partitioned into $M = N/B$ discrete blocks, each of size $B$. 
The sequence $X$ is constructed by concatenating the prompt $P$ with $N$ tokens, denoted as $X = (P, x_1, x_2, \dots, x_N)$, where each token $x_i$ is characterized by a pair $(e_i, z_i)$, comprising continuous embeddings $e_i$ and a token state $z_i$.
Initially, all target positions are initialized as mask tokens, where $z_i = \texttt{[MASK]}$ for all $i \in \{1, \dots, N\}$, and the corresponding embedding sequence is represented as $\mathbf{E} = (e_P, e^{<\text{mask}>}_1, \dots, e^{<\text{mask}>}_N)$. 
During the evolution process, each token $x_i$ transitions through a state space consisting of four distinct stages:
\[
\texttt{[MASK]},\  \mathrm{Soft}([\texttt{MASK}] \cup \mathcal{V}),\  \mathrm{Soft}(\mathcal{V}),\  \texttt{[Decode]},
\]
where $\mathcal{V}$ is the vocabulary.

\begin{figure}[t]
    \centering
    \includegraphics[width=\linewidth]{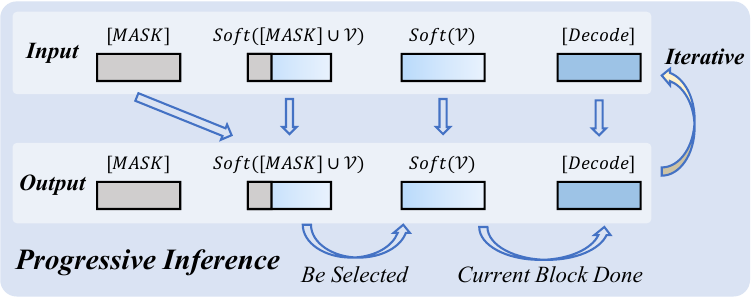}
    \vspace{-8mm}
    \caption{Progressive step-wise token update with blockwise decoding in EvoToken-DLM.}
    \label{fig:inference}
\end{figure}

\paragraph{Token Prediction.} 
At each inference step, we input embeddings $\mathbf{E}$ into the model to obtain a predicted distribution $\{p_i^c\}_{c=1}^{|\mathcal{V}|}$ over the vocabulary for each position $i$.
We retain the top-$K$ probabilities and renormalize them to obtain $\{\hat{p}_i^c\}_{c=1}^{K}$, along with their corresponding tokens $\{\hat{v}_i^c\}_{c=1}^{K} \subseteq \mathcal{V}$.
\textit{Soft embeddings} are then computed as:
\begin{equation}
\begin{aligned}
e_i^{\text{dist}} &= \sum_{c=1}^{K} \hat{p}_i^c \cdot e^{\hat{v}_i^c}, \\
e_i^{\text{dist+M}} &= \alpha \, e_i^{<\text{mask}>} + (1-\alpha) \, e_i^{\text{dist}},
\end{aligned}
\end{equation}
where $\alpha \in [0,1]$ controls the mixing ratio of the mask embedding.

\paragraph{Embedding Assignment by Token State.} 
For token $x_i$, its newly generated embeddings at the current step is assigned based on its current state:
\begin{equation}
e_i =
\begin{cases}
e_i^{<\text{mask}>}, & z_i = \texttt{[MASK]} \\
e_i^{\text{dist+M}}, & z_i = \mathrm{Soft}([\texttt{MASK}] \cup \mathcal{V}) \\
e_i^{\text{dist}}, & z_i = \mathrm{Soft}(\mathcal{V}) \\
e^{v_i}, & z_i = \texttt{[Decode]}
\end{cases}
\end{equation}
where $v_i$ is selected as the token in the vocabulary with the highest confidence among all historical predictions made after $x_i$ enters the $\mathrm{Soft}(\mathcal{V})$ state.

\paragraph{Step-wise Token Update.} 
By default, tokens in the \texttt{[MASK]} state transition to $\mathrm{Soft}([\texttt{MASK}] \cup \mathcal{V})$, whereas tokens already in the $\mathrm{Soft}([\texttt{MASK}] \cup \mathcal{V})$, $\mathrm{Soft}(\mathcal{V})$, or \texttt{[Decode]} states retain their current state. 
At each step, a subset of tokens currently in the \texttt{[MASK]} or $\mathrm{Soft}([\texttt{MASK}] \cup \mathcal{V})$ states in the current block is selected to transition to the $\mathrm{Soft}(\mathcal{V})$ state. Let $S$ denote the set of these selected tokens. 
The complete update rule is formalized as:
\begin{small}
\begin{equation}
z_i \gets
\begin{cases}
\mathrm{Soft}([\texttt{MASK}] \cup \mathcal{V}), & z_i \in \{\mathrm{Soft}([\texttt{MASK}] \cup \mathcal{V}), \\
                                                & \quad \quad \texttt{[MASK]}\} \text{ and } x_i \notin S \\
\mathrm{Soft}(\mathcal{V}), & x_i \in S \text{ or } z_i = \mathrm{Soft}(\mathcal{V}) \\
\texttt{[Decode]}, & z_i = \texttt{[Decode]}
\end{cases}
\end{equation}
\end{small}

\paragraph{Blockwise Decoding.} 
Let $\mathcal{B}$ denote the set of tokens in the current block. Once all tokens in $\mathcal{B}$ reach the $\mathrm{Soft}(\mathcal{V})$ state, they are simultaneously converted to the \texttt{[Decode]} state:
\begin{equation}
\begin{aligned}
z_i &\gets \texttt{[Decode]}, \quad \forall x_i \in \mathcal{B}, \\
    &\text{if all tokens } x_j \in \mathcal{B} \text{ are in the } \mathrm{Soft}(\mathcal{V}) \text{ state}.
\end{aligned}
\end{equation}

As illustrated in Figure~\ref{fig:inference}, combining step-wise token update with blockwise decoding, EvoToken-DLM allows each token to gradually refine its representation from \texttt{[MASK]} to final \texttt{[Decode]} through progressive token evolution.
The detailed algorithm for progressive inference is provided in Appendix~\ref{sec:algorithm-inference}.

\subsection{Continuous Trajectory Supervision}

\begin{figure}[t]
    \centering
    \includegraphics[width=\linewidth]{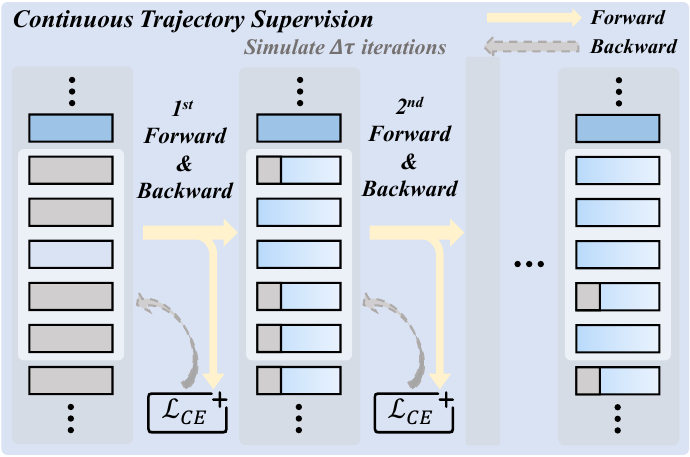}
    \vspace{-8mm}
    \caption{Continuous trajectory supervision by performing $\Delta \tau$ consecutive refinement steps during training and applying supervision at each step, aligning the training objective with the inference process.}
    \label{fig:continuous-sft}
\end{figure}

Unlike conventional masked diffusion frameworks, EvoToken-DLM employs a progressive evolution mechanism. In this approach, the current states and embeddings of the tokens are conditioned on the cumulative history of the preceding refinements. This temporal dependency renders standard single-step denoising objectives inapplicable, necessitating a specialized training paradigm that models the trajectory of token evolution.
We propose continuous trajectory supervision, a training strategy that aligns model optimization with iterative probabilistic token refinement along the diffusion trajectory $\mathcal{T}$, as illustrated in Figure~\ref{fig:continuous-sft}. 
This approach ensures consistency from training to inference.

\paragraph{Initialization and Masking Strategy.}
Given a sequence comprising a prompt and a target response, we sample a contiguous segment of length $L$ from the response as the current training block. To align with the blockwise inference procedure, tokens preceding this block are set to the ground truth, while tokens after this block are replaced with \texttt{[MASK]}. Within the selected block, we randomly mask a subset of tokens to initialize the state $X^{(0)}$.

\paragraph{Trajectory Unrolling.}
Starting from $X^{(0)}$, we simulate $\Delta \tau$ consecutive refinement steps to sample the trajectory:
\begin{small}
\begin{equation}
X^{(i)}, \; \mathcal{L}^{(i)} = \mathrm{Model}(X^{(i-1)}), \quad \forall i = 1, \dots, \Delta \tau,
\end{equation}
\end{small}
where each forward pass produces probability distributions, updated continuous embeddings, and updated token states according to the progressive inference rules described in Section~\ref{sec:progressive_inference}. 

\paragraph{Cumulative Trajectory Loss.}
At each step $i$, we compute a supervised loss $\mathcal{L}^{(i)}$ against the ground-truth tokens within the current block. Rather than backpropagating only through the final step, we perform a backward pass for every forward step:
\begin{equation}
\nabla_\theta \mathcal{L}^{(i)}, \quad i = 1, \dots, \Delta \tau.
\end{equation}
By explicitly simulating the progressive refinement during training, continuous trajectory supervision aligns the learning objective with the inference behavior of EvoToken-DLM. 
The detailed algorithm for continuous trajectory supervision is provided in Appendix~\ref{sec:algorithm-sft}.

\subsection{Extension to Blockwise Diffusion}



EvoToken-DLM naturally extends to blockwise diffusion by partitioning the sequence into consecutive blocks. Within each block, tokens undergo full progressive refinement before the generation moves to the next, preserving the global autoregressive structure while enabling intra-block parallelism.
For training, we adapt continuous trajectory supervision to this setting. Following existing frameworks~\cite{d2f, sdar, llada2}, we exploit block-level causal dependencies to enable independent, parallel training of blocks. Within each block, the continuous trajectory supervision procedure simulates $\Delta \tau$ refinement steps for supervision.
\section{Experiments}

\subsection{Experimental Setup}
\label{sec:experimental-setup}

We employ LLaDA-Instruct-8B~\citep{llada} as our primary backbone for fine-tuning. To evaluate cross-model consistency, we also apply our method to LLaDA-1.5~\citep{llada1.5}, Dream-Instruct-7B~\citep{dream2025} and D2F-LLaDA~\citep{d2f}, the last of which serves as the base model for our blockwise diffusion experiments.
For fine-tuning, we utilize the S1K dataset~\citep{dataset-s1k} and train the pretrained model for a default duration of 10k steps using continuous trajectory supervision.
Evaluations are performed across several mathematical and reasoning benchmarks, including Countdown~\citep{dataset-countdown}, GSM8K~\citep{dataset-gsm8k}, MATH500~\citep{dataset-math500}, and SVAMP~\citep{dataset-svamp}.
More details are presented in Appendix~\ref{sec:more-implementation-details}.

\subsection{Evaluation Results}

\begin{table*}[t] 
\centering
\caption{Performance comparison on the Countdown, GSM8K, MATH500 and SVAMP datasets across various generation lengths and NFEs based on LLaDA-Instruct-8B. EvoToken-DLM is initialized from LLaDA-Instruct-8B and fine-tuned for 10k steps using continuous trajectory supervision. Comparisons are conducted against both the baseline model and the sft-baseline.}
\label{tab:main-results} 
\vspace{-4mm}
\setlength{\tabcolsep}{3.5pt} 
\resizebox{\textwidth}{!}{
    \begin{tabular}{ll|cccc|cccc|cccc|cccc}
        \toprule
        &
        & \multicolumn{4}{c|}{\textbf{Countdown}}
        & \multicolumn{4}{c|}{\textbf{GSM8K}} 
        & \multicolumn{4}{c|}{\textbf{MATH500}} 
        & \multicolumn{4}{c}{\textbf{SVAMP}} \\
        \cmidrule(lr){3-6} \cmidrule(lr){7-10} \cmidrule(lr){11-14} \cmidrule(lr){15-18}
        $\boldsymbol{\frac{NFE}{Gen\ Len}}$ & \textbf{Method} & 128 & 256 & 512 & \cellcolor{gray!10}Avg. & 128 & 256 & 512 & \cellcolor{gray!10}Avg. & 128 & 256 & 512 & \cellcolor{gray!10}Avg. & 128 & 256 & 512 & \cellcolor{gray!10}Avg. \\
        \midrule
        \multirow{5}{*}{\textbf{1}} 
        & Baseline & 21.48 & 23.83 & 20.70 & \cellcolor{gray!10}22.00 & 70.20 & 79.30 & 83.47 & \cellcolor{gray!10}77.66 & 28.80 & 34.60 & 39.40 & \cellcolor{gray!10}34.27 & 88.33 & 84.67 & 86.00 & \cellcolor{gray!10}86.33 \\
        \cmidrule(lr){2-18}
        & FT-Base (10k FT) & 33.20 & 21.48 & 19.53 & \cellcolor{gray!10}24.74 & 71.04 & 82.11 & 82.56 & \cellcolor{gray!10}78.57 & 26.60 & 36.20 & 40.40 & \cellcolor{gray!10}34.40 & 87.67 & 89.67 & 89.67 & \cellcolor{gray!10}89.00 \\
        \cmidrule(lr){2-18}
        & \multirow{2}{*}{\textbf{EvoToken (10k FT)}} & 39.84 & 35.55 & 42.97 & \cellcolor{gray!10}\textbf{39.45} & 74.30 & 83.47 & 84.46 & \cellcolor{gray!10}\textbf{80.74} & 28.40 & 39.60 & 41.00 & \cellcolor{gray!10}\textbf{36.33} & 89.00 & 89.67 & 90.00 & \cellcolor{gray!10}\textbf{89.56} \\
        &  & \textcolor{green!70!black}{+18.36} & \textcolor{green!70!black}{+11.72} & \textcolor{green!70!black}{+22.27} & \cellcolor{gray!10}\textcolor{green!70!black}{+17.45} & \textcolor{green!70!black}{+4.10} & \textcolor{green!70!black}{+4.17} & \textcolor{green!70!black}{+0.99} & \cellcolor{gray!10}\textcolor{green!70!black}{+3.08} & \textcolor{gray}{-0.40} & \textcolor{green!70!black}{+5.00} & \textcolor{green!70!black}{+1.60} & \cellcolor{gray!10}\textcolor{green!70!black}{+2.06} & \textcolor{green!70!black}{+0.67} & \textcolor{green!70!black}{+5.00} & \textcolor{green!70!black}{+4.00} & \cellcolor{gray!10}\textcolor{green!70!black}{+3.23} \\

        \midrule
        \midrule
        \multirow{5}{*}{\textbf{$\boldsymbol{\frac{1}{2}}$}} 
        & Baseline & 26.17 & 16.41 & 16.80 & \cellcolor{gray!10}19.79 & 67.55 & 77.63 & 79.83 & \cellcolor{gray!10}75.00 & 26.60 & 32.20 & 33.20 & \cellcolor{gray!10}30.67 & 86.00 & 86.67 & 84.00 & \cellcolor{gray!10}85.56 \\
        \cmidrule(lr){2-18}
        & FT-Base (10k FT) & 28.12 & 16.80 & 16.41 & \cellcolor{gray!10}20.44 & 63.91 & 78.62 & 79.00 & \cellcolor{gray!10}73.84 & 22.60 & 31.20 & 34.00 & \cellcolor{gray!10}29.27 & 85.00 & 87.00 & 89.33 & \cellcolor{gray!10}87.11 \\
        \cmidrule(lr){2-18}
        & \multirow{2}{*}{\textbf{EvoToken (10k FT)}} & 34.77 & 30.08 & 30.08 & \cellcolor{gray!10}\textbf{31.64} & 73.54 & 82.03 & 81.80 & \cellcolor{gray!10}\textbf{79.12} & 29.20 & 36.40 & 37.40 & \cellcolor{gray!10}\textbf{34.33} & 89.33 & 92.33 & 89.67 & \cellcolor{gray!10}\textbf{90.44} \\
        &  & \textcolor{green!70!black}{+8.60} & \textcolor{green!70!black}{+13.67} & \textcolor{green!70!black}{+13.28} & \cellcolor{gray!10}\textcolor{green!70!black}{+11.85} & \textcolor{green!70!black}{+5.99} & \textcolor{green!70!black}{+4.40} & \textcolor{green!70!black}{+1.97} & \cellcolor{gray!10}\textcolor{green!70!black}{+4.12} & \textcolor{green!70!black}{+2.60} & \textcolor{green!70!black}{+4.20} & \textcolor{green!70!black}{+4.20} & \cellcolor{gray!10}\textcolor{green!70!black}{+3.66} & \textcolor{green!70!black}{+3.33} & \textcolor{green!70!black}{+5.66} & \textcolor{green!70!black}{+5.67} & \cellcolor{gray!10}\textcolor{green!70!black}{+4.88} \\
    
        \midrule
        \midrule
        \multirow{5}{*}{\textbf{$\boldsymbol{\frac{1}{4}}$}} 
        & Baseline & 17.19 & 15.62 & 16.41 & \cellcolor{gray!10}16.41 & 59.14 & 68.23 & 66.57 & \cellcolor{gray!10}64.65 & 23.40 & 26.60 & 29.60 & \cellcolor{gray!10}26.53 & 81.00 & 77.33 & 75.00 & \cellcolor{gray!10}77.78 \\
        \cmidrule(lr){2-18}
        & FT-Base (10k FT) & 14.06 & 13.67 & 9.77 & \cellcolor{gray!10}12.50 & 49.05 & 62.17 & 61.87 & \cellcolor{gray!10}57.70 & 16.20 & 19.60 & 23.20 & \cellcolor{gray!10}19.67 & 66.67 & 75.33 & 72.00 & \cellcolor{gray!10}71.33 \\
        \cmidrule(lr){2-18}
        & \multirow{2}{*}{\textbf{EvoToken (10k FT)}} & 23.05 & 16.02 & 12.11 & \cellcolor{gray!10}\textbf{17.06} & 64.82 & 75.74 & 72.33 & \cellcolor{gray!10}\textbf{70.96} & 23.60 & 31.00 & 31.20 & \cellcolor{gray!10}\textbf{28.60} & 78.33 & 83.33 & 81.33 & \cellcolor{gray!10}\textbf{81.00} \\
        &  & \textcolor{green!70!black}{+5.86} & \textcolor{green!70!black}{+0.40} & \textcolor{gray}{-4.30} & \cellcolor{gray!10}\textcolor{green!70!black}{+0.65} & \textcolor{green!70!black}{+5.68} & \textcolor{green!70!black}{+7.51} & \textcolor{green!70!black}{+5.76} & \cellcolor{gray!10}\textcolor{green!70!black}{+6.31} & \textcolor{green!70!black}{+0.20} & \textcolor{green!70!black}{+4.40} & \textcolor{green!70!black}{+1.60} & \cellcolor{gray!10}\textcolor{green!70!black}{+2.07} & \textcolor{gray}{-2.67} & \textcolor{green!70!black}{+6.00} & \textcolor{green!70!black}{+6.33} & \cellcolor{gray!10}\textcolor{green!70!black}{+3.22} \\
        \bottomrule
    \end{tabular}
}
\end{table*}

\begin{figure}[t]
    \centering
    \includegraphics[width=\linewidth]{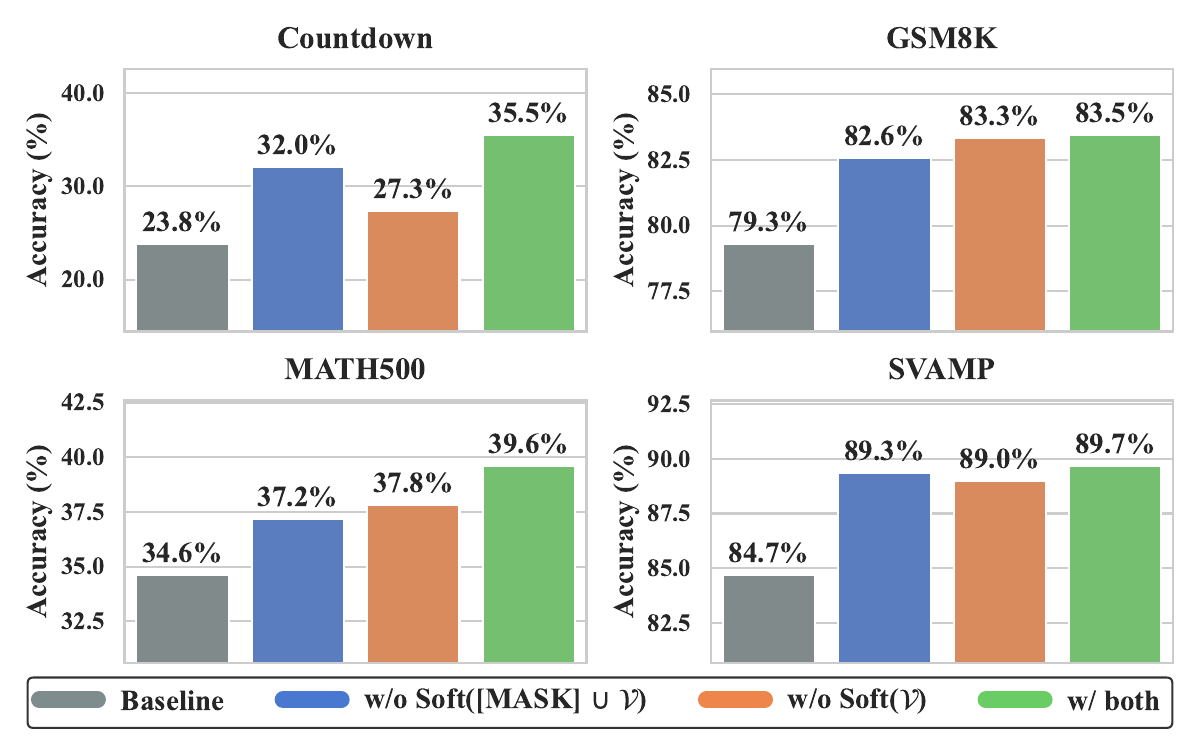}
    \vspace{-8mm}
    \caption{Ablation study on the presence of intermediate refinement states in EvoToken-DLM.}
    \label{fig:ablation-states}
\end{figure}

\begin{figure*}[t]
    \centering
    \includegraphics[width=\linewidth]{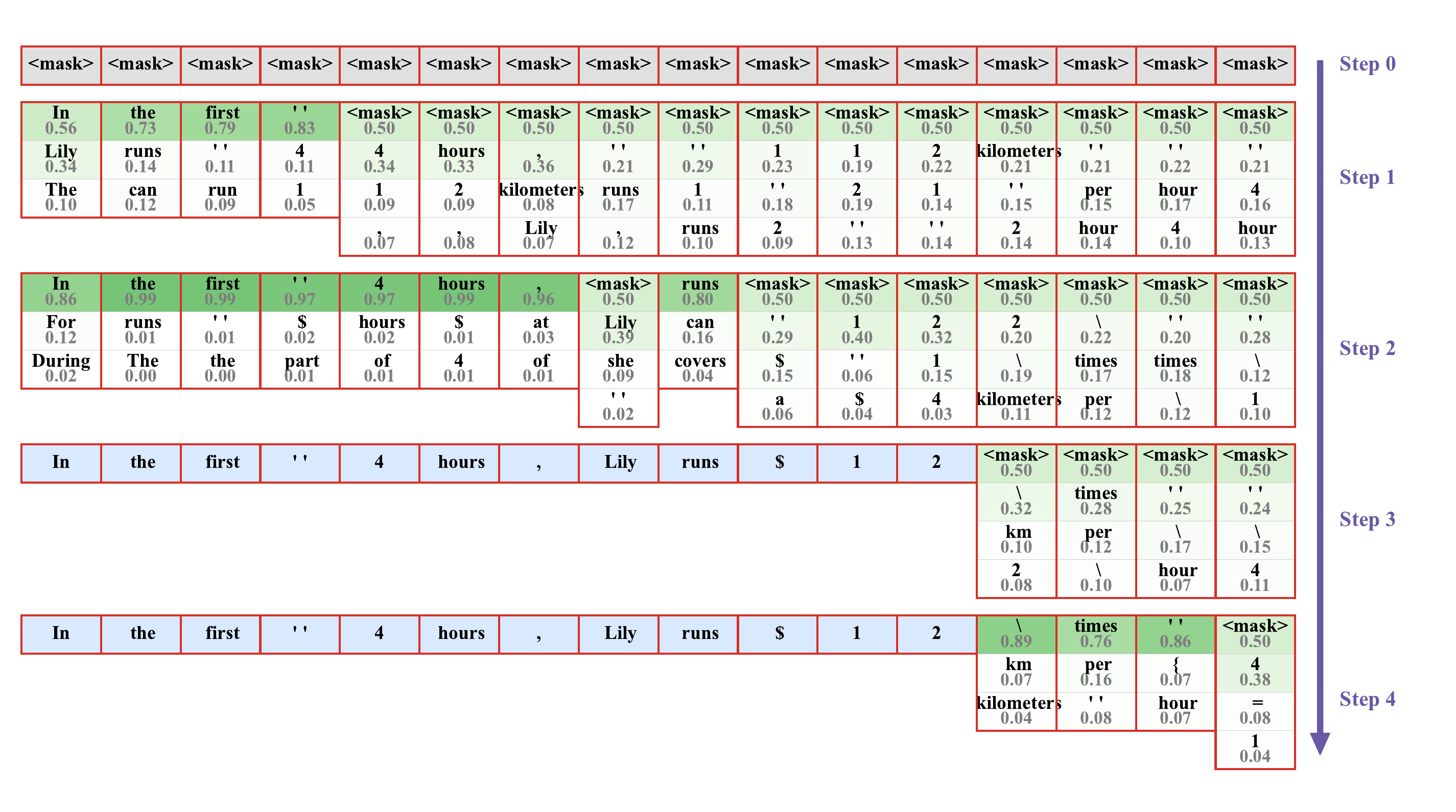}
    \vspace{-8mm}
    \caption{
    An illustrative example of EvoToken-DLM during inference, showing intermediate refinement states for a selected subsequence across successive steps.
    The block size is set to 12, and the refinement process for the first 16 output tokens is visualized. For each position, only the top $K=3$ most probable tokens are retained.
    }
    \label{fig:refine-example}
\end{figure*}

\begin{figure}[t]
    \centering
    \includegraphics[width=\linewidth]{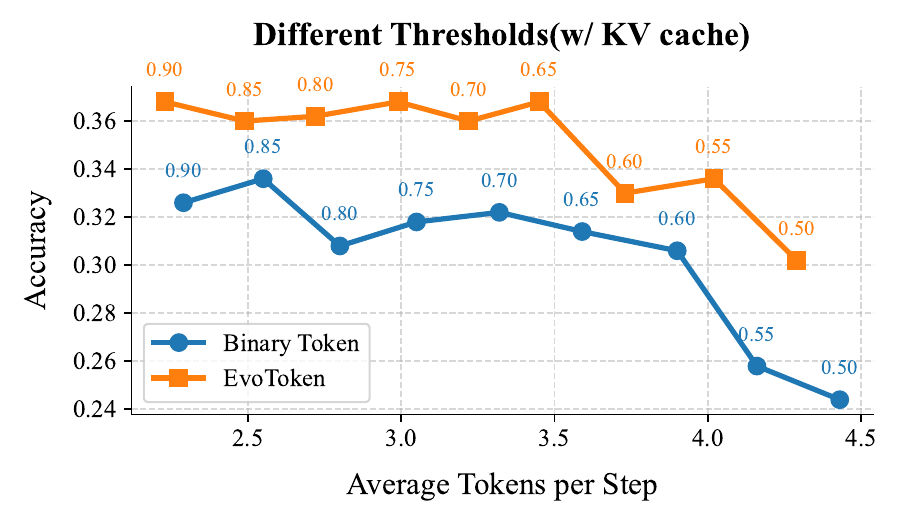}
    \vspace{-8mm}
    \caption{Comparison between EvoToken and binary masking baseline on MATH500 with KV-caching and different confidence thresholds. EvoToken consistently achieves higher accuracy than baseline across various thresholds under the same average tokens per step.}
    \label{fig:threshold-comparison}
\end{figure}


\begin{figure}[t]
    \centering
    \includegraphics[width=\linewidth]{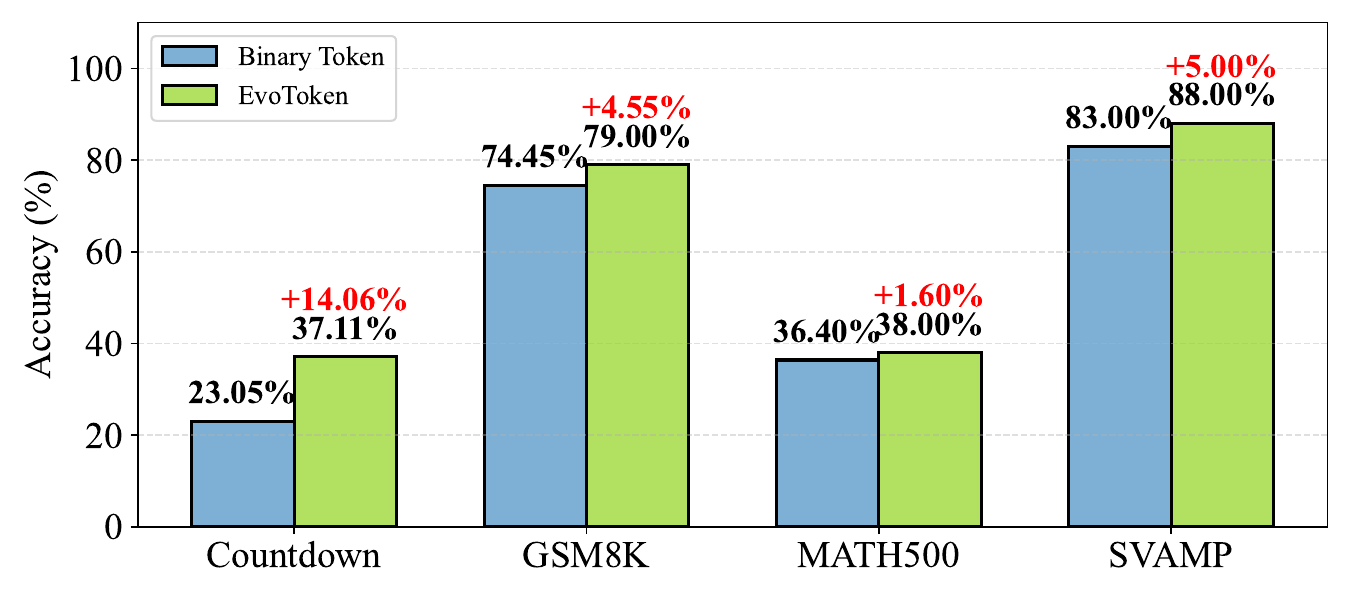}
    \vspace{-8mm}
    \caption{Comparison between EvoToken and the binary masking baseline based on another pretrained model Dream-Instruct-7B. We apply continuous trajectory supervision and evaluate performance on various datasets.}
    \label{fig:dream-results}
\end{figure}

\begin{figure}[t]
    \centering
    \includegraphics[width=\linewidth]{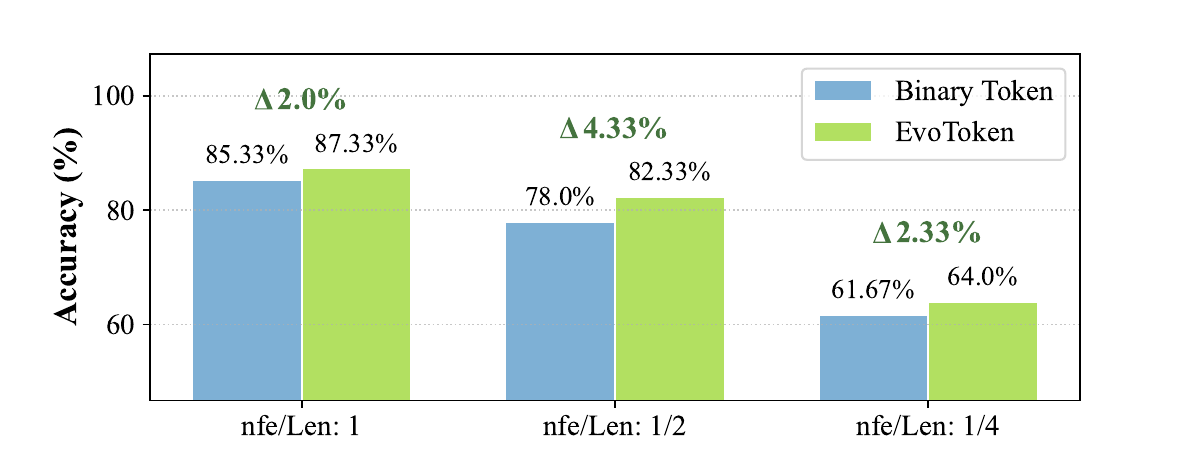}
    \vspace{-8mm}
    \caption{Comparison between EvoToken and the binary masking baseline based on blockwise diffusion model D2F-LLaDA. We apply continuous trajectory supervision and evaluate performance on SVAMP.}
    \label{fig:d2f-results}
\end{figure}

\begin{figure}[t]
    \centering
    \includegraphics[width=\linewidth]{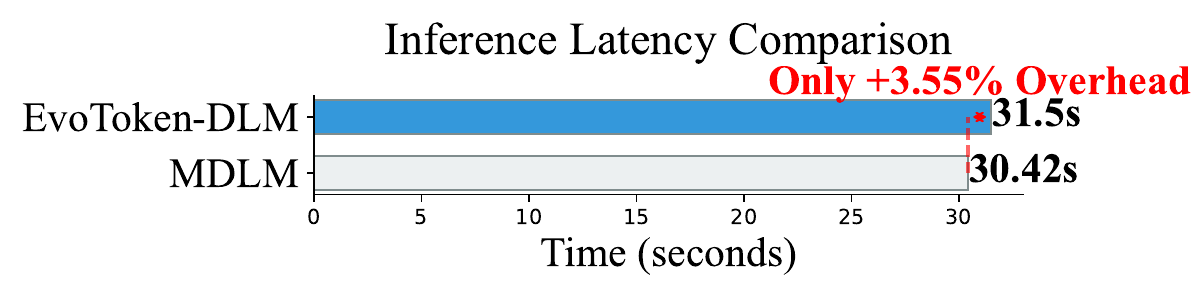}
    \vspace{-8mm}
    \caption{EvoToken-DLM exhibits competitive inference efficiency, introducing minimal latency penalties relative to standard MDLM architectures.}
    \label{fig:time_cost_comparison}
\end{figure}

\begin{figure}[t]
    \centering
    \includegraphics[width=\linewidth]{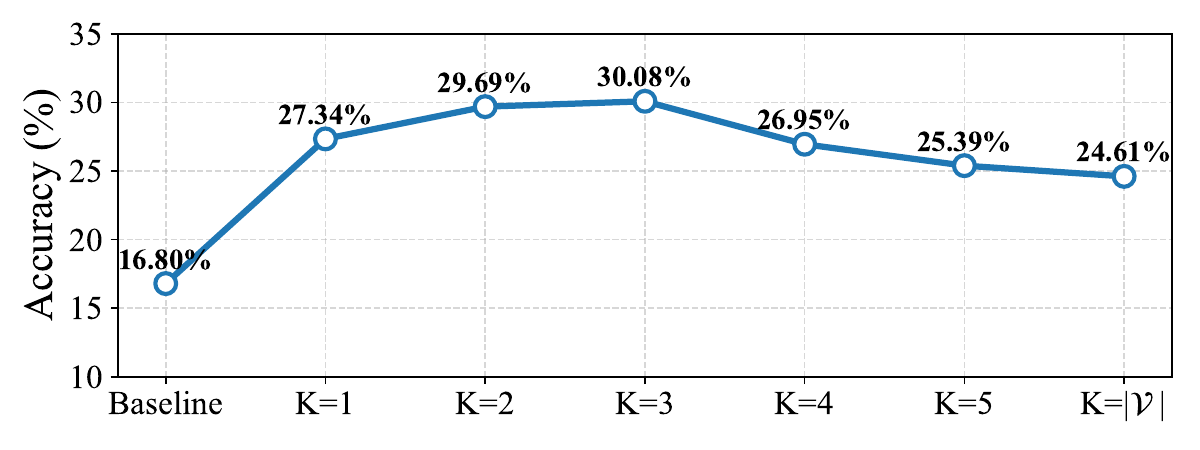}
    \vspace{-8mm}
    \caption{Performance comparison between the baseline and EvoToken-DLM with different top-$K$ settings on Countdown.}
    \label{fig:topk-comparison}
\end{figure}

\paragraph{Main Performance Comparison.} 
Table~\ref{tab:main-results} compares EvoToken-DLM against the original LLaDA-Instruct-8B and the FT-baseline across multiple reasoning benchmarks. 
EvoToken-DLM predominantly surpasses both baselines, exhibiting substantial performance leaps under varying configurations. 
Specifically, at $\frac{NFE}{Gen\  Len}=1$, our method yields average accuracy gains of $17.45\%$ on Countdown, $3.08\%$ on GSM8K, $2.06\%$ on MATH500, and $3.23\%$ on SVAMP compared to the original model. 
These results underscore the superiority of our soft token evolution framework in enhancing reasoning capabilities and generation quality.
Additional results with different block sizes are presented in Appendix~\ref{sec:additional-results-different-block-size}, and qualitative comparisons are presented in Appendix~\ref{sec:qualitative-comparisons}.

\paragraph{Importance of Intermediate States.}
Figure~\ref{fig:ablation-states} illustrates the ablation study on the presence of intermediate refinement states. The performance drop observed when removing these states confirms that the gradual transition from mask to soft-token is essential for the model to iteratively refine its predictions.

\paragraph{Qualitative Visualization.}
We provide a qualitative visualization of the inference process in Figure~\ref{fig:refine-example}. By tracing the evolution of a selected subsequence, we observe how initial uncertain tokens progressively converge into precise and coherent results. This visualization confirms that EvoToken-DLM effectively implements a progressive refinement mechanism, allowing the model to iteratively calibrate its predictions within the diffusion framework.
Additional qualitative visualization results of the inference process are presented in Appendix~\ref{sec:additional-inference-examples}.

\paragraph{Compatibility with KV-Caching.} 
To further demonstrate the practical efficiency of EvoToken-DLM, we integrate it with the KV-caching mechanism as proposed in Fast-dLLM~\citep{wu2025fast}. 
This integration is essential to ensure that our adaptive token evolution does not interfere with the accelerated inference pipelines of DLMs. 
As reported in Table~\ref{tab:with-kv-cache}, we evaluate the performance of EvoToken-DLM equipped with KV-caching against the baseline on Countdown. The results indicate that our method consistently maintains superior performance with KV-caching across various computational budgets, proving its seamless integration with the KV-caching mechanism.

\paragraph{Robustness across Thresholds.}
We further adopt another parallel generation strategy using the confidence threshold proposed in Fast-dLLM~\citep{wu2025fast} to replace the fixed NFE setting. This allows for a more flexible allocation of computational resources during inference. 
We analyze the sensitivity of our method to different thresholds on the MATH500 dataset, with KV-caching enabled. 
As illustrated in Figure~\ref{fig:threshold-comparison}, EvoToken consistently outperforms the binary masking baseline given the same average token budget per step.
These results demonstrate the superior adaptability of EvoToken-DLM.

\begin{table}[t]
\centering
\caption{Performance comparison on Countdown with KV-caching. EvoToken integrates seamlessly with KV-caching mechanism.}
\label{tab:with-kv-cache}
\resizebox{\columnwidth}{!}{
    \begin{tabular}{cc|ll}
        \toprule
        \textbf{Gen Len} & $\boldsymbol{\frac{NFE}{Gen\ Len}}$ & \textbf{Baseline+Cache} & \textbf{EvoToken+Cache} \\ 
        \midrule
        \multirow{3}{*}{128} & 1   & 21.88 & \textbf{36.33} \small{\textcolor{green!70!black}{(+14.45)}} \\ 
                             & 1/2 & 23.05 & \textbf{28.52} \small{\textcolor{green!70!black}{(+5.47)}} \\
                             & 1/4 & 7.42  & \textbf{18.75} \small{\textcolor{green!70!black}{(+11.33)}} \\
        \cmidrule(lr){1-4}
        \multirow{3}{*}{256} & 1   & 21.48 & \textbf{28.52} \small{\textcolor{green!70!black}{(+7.04)}} \\
                             & 1/2 & 19.92 & \textbf{25.00} \small{\textcolor{green!70!black}{(+5.08)}} \\
                             & 1/4 & 9.77  & \textbf{12.50} \small{\textcolor{green!70!black}{(+2.73)}} \\ 
        \bottomrule
    \end{tabular}
}
\end{table}

\paragraph{Generalization Across Models.}

We evaluate the transferability of our approach by applying continuous trajectory supervision to the Dream-Instruct-7B pretrained base. As shown in Figure~\ref{fig:dream-results}, the improvements observed in the primary model consistently generalize to the alternative backbone. This consistency underscores that EvoToken-DLM serves as a general enhancement for diffusion language models.
Additional results based on LLaDA-1.5 are presented in Appendix~\ref{sec:additional-results-llada-1.5}.

\paragraph{Extension to Blockwise Diffusion.}
To further validate the versatility of EvoToken, we extend our method to the blockwise diffusion framework, specifically using D2F-LLaDA as the base model. As illustrated in Figure~\ref{fig:d2f-results}, EvoToken significantly outperforms the binary masking baseline, proving its robustness and adaptability.

\paragraph{Inference Efficiency.}
Figure~\ref{fig:time_cost_comparison} illustrates that EvoToken-DLM introduces only negligible latency compared to standard MDLMs, with the marginal overhead stemming primarily from the element-wise addition of token embeddings during the refinement process.
Such minimal overhead, alongside substantial improvements, makes EvoToken-DLM highly practical for real-world deployment.

\paragraph{Impact of Top-$K$ Filtering.} 
In Figure~\ref{fig:topk-comparison}, we analyze the sensitivity of the model to different top-$K$ settings during the  refinement process. EvoToken-DLM shows robust performance across a wide range of $K$ values, consistently outperforming the baseline.
An additional ablation study on the mixing ratio $\alpha$ is presented in Appendix~\ref{sec:ablation-mixing-ratio}.

\noindent More analyses regarding rapid adaptation from pretrained MDLMs are presented in Appendix~\ref{sec:more-analyses}.

\section{Related Work}

\subsection{MDLMs}

Masked Diffusion Language Models (MDLMs) adapt the diffusion paradigm~\citep{DDPM, podell2023sdxl, rombach2022high, song2019generative, DDIM} to discrete text generation. Building on foundational work in noise scheduling and objectives~\citep{d3pm, mdlm, llada2, yang2025mmada, llada, wu2025dmark, wang2025time}, recent large-scale models like LLaDA~\citep{llada} and Dream~\citep{dream2025} have shown that MDLMs can match autoregressive baselines in complex reasoning. 
Despite their potential, the iterative denoising process remains computationally expensive. Current research addresses this through two main efficiency frontiers: developing specialized caching mechanisms and architecting blockwise generative processes.

\paragraph{KV-Cache Optimization for MDLMs.}
Standard KV caching is incompatible with bidirectional MDLMs, often necessitating recomputation per step. Fast-dLLM~\citep{wu2025fast, wu2025fast2} mitigates this via block-wise approximate caching, while others~\citep{ma2025dinfer, shen2025improving, bao2025learning} refine generation coherence. Furthermore, dKV-Cache~\citep{ma2025dkv} and dLLM-Cache~\citep{liu2025dllm} utilize selective token updates, and Sparse-dLLM~\citep{song2025sparse} applies dynamic eviction to reduce long-context memory overhead.

\paragraph{Blockwise Diffusion Language Models.}
Blockwise MDLMs~\citep{han2023ssd, arriolablock, zhao2025d1, liu2025longllada, liu2025sequential, sdar, d2f} hybridize AR global ordering with intra-block diffusion to support KV-caching. 
To eliminate serial bottlenecks, \citet{d2f} introduces D2F, which enables decoding future blocks from noisy intermediate states.

\subsection{Latent Reasoning}

\paragraph{Reasoning in Continuous Space.} To enhance Chain-of-Thought expressivity, recent works transition from discrete tokens to latent spaces. \citet{hao2024training} leverage transformer hidden states, while \citet{xu2025softcot, xu2025softcotpp} and \citet{zhang2025soft, zhuangmixture} utilize projection modules or probability-weighted embeddings. 


\paragraph{Latent Reasoning with DLMs.} To mitigate remasking information loss, recent DLMs integrate continuous semantics. \citet{hersche2025soft} propose Soft Masking via dynamic embedding blends, while \citet{zheng2025continuously} employ dual discrete-continuous diffusion. Additionally, \citet{kang2025ladir} utilize VAE-based latent spaces to refine reasoning trajectories. 
HDLM \citep{zhou2025next} further structures this process through a hierarchical vocabulary for coarse-to-fine refinement, whereas CCDD \citep{zhou2025coevolutionary} co-evolves continuous and discrete modalities in a joint space to balance expressivity with trainability.
\section{Conclusion}


In this paper, we presented EvoToken-DLM, a novel diffusion language modeling approach that replaces rigid binary masks with evolving soft token distributions. This shift enables a progressive decoding process, overcoming the limitations of irreversible discrete assignments in traditional MDLMs. By introducing continuous trajectory supervision, we effectively align the training objective with iterative probabilistic refinement. Extensive experiments demonstrate that EvoToken-DLM consistently outperforms strong baselines while remaining fully compatible with KV-caching and blockwise architectures.

\section*{Limitations}

While our approach enables rapid adaptation from pretrained MDLMs to EvoToken-DLM via lightweight supervised fine-tuning, it faces training challenges when applied to models initialized with autoregressive (AR) priors. The inherent discrepancy between unidirectional AR pretraining and our iterative bidirectional refinement process leads to increased training difficulty and slower convergence for AR-based backbones. We provide a detailed comparative analysis in Appendix~\ref{sec:more-analyses}.


\bibliography{main}

\clearpage
\appendix
\renewcommand\thesection{\Alph{section}}
\renewcommand\thefigure{S\arabic{figure}}
\renewcommand\thetable{S\arabic{table}}
\renewcommand\theequation{S\arabic{equation}}
\renewcommand\thealgorithm{S\arabic{algorithm}}

\renewcommand\theHfigure{S\arabic{figure}}
\renewcommand\theHtable{S\arabic{table}}
\renewcommand\theHequation{S\arabic{equation}}
\renewcommand\theHalgorithm{S\arabic{algorithm}}

\setcounter{figure}{0}
\setcounter{table}{0}
\setcounter{equation}{0}
\setcounter{algorithm}{0}



\twocolumn[

\section*{Appendix}

\section{Appendix Overview}

\begin{figure*}[t]

    \subsection{Additional Inference Examples of EvoToken-DLM}
    \label{sec:additional-inference-examples}
    
    We provide additional detailed visualizations of the internal refinement process to demonstrate how EvoToken-DLM iteratively refines intermediate states. 
    As shown in Figure~\ref{fig:refine-example-additional-1} and Figure~\ref{fig:refine-example-additional-2}, for the prompt involving arithmetic, the model progressively clarifies the soft token representations. The visualization highlights how uncertain embeddings at early simulation steps are refined into sharp, symbolically correct tokens as the step increases.
        
    \centering
    \includegraphics[width=\linewidth]{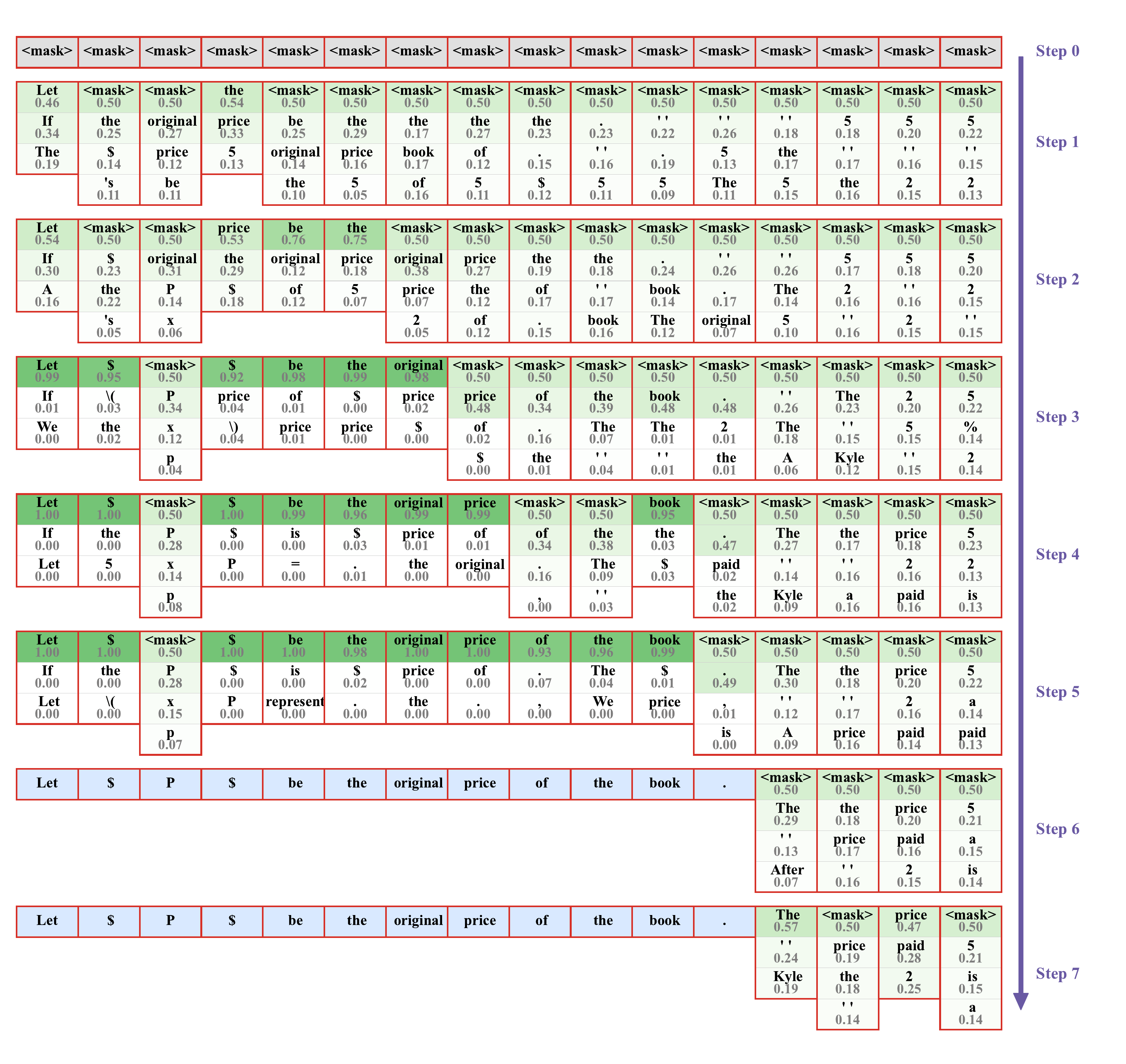}
    \vspace{-8mm}
    \caption{
    Additional inference example with a block size of 12, showing intermediate refinement states for a selected subsequence across successive steps. We showcase the refinement states for the first 16 output tokens based on the prompt: "Kyle bought last year's best-selling book for \$19.50. This is with a 25\% discount from the original price. What was the original price of the book?".
    }
    \label{fig:refine-example-additional-1}
\end{figure*}

\begin{itemize}
    \item \textbf{Appendix~\ref{sec:more-method-details}: More Methodological Details}
    \begin{itemize}
        \item Appendix~\ref{sec:algorithm-inference}: EvoToken Algorithm for Progressive Inference
        \item Appendix~\ref{sec:algorithm-sft}: EvoToken Algorithm for Continuous Trajectory Supervision
    \end{itemize}
    
    \item \textbf{Appendix~\ref{sec:more-implementation-details}: More Implementation Details}
    \begin{itemize}
        \item Appendix~\ref{sec:more-implementation-details-dataset}: Detailed Descriptions of Training Dataset and Evaluation Benchmarks
        \item Appendix~\ref{sec:more-implementation-details-continuous-sft}: Training Configurations for Continuous Trajectory Supervision
        \item Appendix~\ref{sec:more-implementation-details-inference-eval}: Inference and Evaluation Setup
    \end{itemize}

    \item \textbf{Appendix~\ref{sec:more-analyses}: More Analyses}
    \begin{itemize}
        \item Appendix~\ref{sec:easy-adaptation-mdlm}: Paradigm Consistency: Why MDLMs Rapidly Adapt to EvoToken-DLM
        \item Appendix~\ref{sec:hard-adaptation-ar}: Adaptation Hurdles for AR Backbones: Causal Prior Mismatch
    \end{itemize}

    \item \textbf{Appendix~\ref{sec:more-experimental-results}: More Experimental Results}
    \begin{itemize}
        \item Appendix~\ref{sec:additional-results-llada-1.5}: Additional Results Based on LLaDA-1.5
        \item Appendix~\ref{sec:ablation-mixing-ratio}: Ablation Study on the Mixing Ratio $\alpha$
        \item Appendix~\ref{sec:additional-results-different-block-size}: Additional Results with Different Block Sizes
        \item Appendix~\ref{sec:additional-inference-examples}: Additional Inference Examples of EvoToken-DLM
        \item Appendix~\ref{sec:qualitative-comparisons}: Qualitative Comparisons for EvoToken-DLM and MDLMs
    \end{itemize}

\end{itemize}

]

\section{More Methodological Details}
\label{sec:more-method-details}

\subsection{EvoToken Algorithm for Progressive Inference}
\label{sec:algorithm-inference}

\begin{algorithm*}[ht]
\caption{Progressive Inference with EvoToken-DLM}
\label{alg:progressive-inference}
\begin{algorithmic}[1]
\Require Prompt $P$, target length $N$, block size $B$, mixing ratio $\alpha$, filtering threshold $K$
\Ensure Decoded sequence $V = (v_1, \dots, v_N)$
\State Initialize states $\mathbf{Z} = (z_1, \dots, z_N)$ where $z_i \gets \texttt{[MASK]}, \quad \forall i \in \{1, \dots, N\}$
\State Initialize embeddings $\mathbf{E} = (e_P, e_1, \dots, e_N)$ where $e_i \gets e^{<\text{mask}>}, \quad \forall i \in \{1, \dots, N\}$
\For{block $b = 1 \to N/B$}
    \State $\mathcal{B}_b \gets$ indices of the current block
    \While{$\exists i \in \mathcal{B}_b$ s.t. $z_i \neq \texttt{[Decode]}$}
        \State $\{v_i^c, p_i^c\}_{c=1}^{|\mathcal{V}|} \gets \text{Model}(\mathbf{E})$
        \State $\{\hat{v}_i^c, \hat{p}_i^c\}_{c=1}^{K} \gets \text{Normalize}(\text{TopK}(\{v_i^c, p_i^c\}_{c=1}^{|\mathcal{V}|}, K))$
        \For{$i \in \{1, \dots, N\}$}
            \State $e_i^{\text{dist}} \gets \sum_{c=1}^{K} \hat{p}_i^c \cdot e^{\hat{v}_i^c}$
            \State $e_i^{\text{dist+M}} \gets \alpha e^{<\text{mask}>} + (1-\alpha) e_i^{\text{dist}}$
        \EndFor
        \State Select subset $S \subseteq \{i \in \mathcal{B}_b \mid z_i \in \{\texttt{[MASK]}, \mathrm{Soft}([\texttt{MASK}] \cup \mathcal{V})\} \}$
        \State $z_i \gets \mathrm{Soft}(\mathcal{V}), \quad \forall i \in S$
        \State $z_i \gets \mathrm{Soft}([\texttt{MASK}] \cup \mathcal{V}), \quad \forall i \notin S \text{ s.t. } z_i = \texttt{[MASK]}$
        \If{$\forall i \in \mathcal{B}_b, z_i = \mathrm{Soft}(\mathcal{V})$}
            \State $z_i \gets \texttt{[Decode]}, \quad \forall i \in \mathcal{B}_b$
            \State Identify $v_i$ as the highest-confidence token since $z_i$ transitioned to $\mathrm{Soft}(\mathcal{V}), \quad \forall i \in \mathcal{B}_b$
        \EndIf
        \For{$i \in \{1, \dots, N\}$}
            \State $e_i \gets 
            \begin{cases} 
            e^{<\text{mask}>} & \text{if } z_i = \texttt{[MASK]} \\
            e_i^{\text{dist+M}} & \text{if } z_i = \mathrm{Soft}([\texttt{MASK}] \cup \mathcal{V}) \\
            e_i^{\text{dist}} & \text{if } z_i = \mathrm{Soft}(\mathcal{V}) \\
            e^{v_i} & \text{if } z_i = \texttt{[Decode]} 
            \end{cases}$
        \EndFor
    \EndWhile
\EndFor
\State \Return $V = (v_1, \dots, v_N)$
\end{algorithmic}
\end{algorithm*}

As presented in Algorithm~\ref{alg:progressive-inference}, the core of the inference process lies in the management of the token states $\mathbf{Z}$ and continuous embeddings $\mathbf{E}$. Each token position starts in the \texttt{[MASK]} state. For each block $\mathcal{B}_b$, the model performs multiple forward passes to refine the soft embeddings. In each step, tokens in a subset $S$ are promoted from mask or soft-mask states to pure soft states. To ensure the stability of the final output, we track the historical high-confidence predictions $v_i$ for each position since it entered the soft state. Once all tokens in the current block are in pure soft states, they are transitioned to the \texttt{[Decode]} state, and their hard embeddings are used as context for the next block. This mechanism effectively facilitates a progressive refinement process for each token.

\subsection{EvoToken Algorithm for Continuous Trajectory Supervision}
\label{sec:algorithm-sft}

\begin{algorithm*}[ht]
\caption{Continuous Trajectory Supervision for EvoToken-DLM}
\label{alg:continuous-sft}
\begin{algorithmic}[1]
\Require Training dataset $\mathcal{D}$, refinement steps $T$, learning rate $\eta$, total iterations $N_{iter}$
\Ensure Optimized parameters $\theta$
\For{iteration $n = 1 \to N_{iter}$}
    \State Sample training pair $(X, Y) \sim \mathcal{D}$
    \State Sample a target block $\mathcal{B} \subseteq Y$
    \State Partition $\mathcal{B}$ into subset $S_{soft}$ and $S_{mask}$
    \State Initialize states $\mathbf{Z}^{(0)}$:
    \State \quad $z_i \gets \texttt{[Decode]}, \quad \forall i < \mathcal{B}$
    \State \quad $z_i \gets \mathrm{Soft}(\mathcal{V}), \quad \forall i \in S_{soft}$
    \State \quad $z_i \gets \texttt{[MASK]}, \quad \forall i \in S_{mask} \cup \{i > \mathcal{B}\}$
    \State Initial embeddings $\mathbf{E}^{(0)}$:
    \State \quad $e_X^{(0)} \gets e^{\{X\}}$
    \State \quad $e_i^{(0)} \gets e^{y_i}, \quad \forall i \in S_{soft} \cup \{i < \mathcal{B}\}$
    \State \quad $e_i^{(0)} \gets e^{<\text{mask}>}, \quad \forall i \in S_{mask} \cup \{i > \mathcal{B}\}$
    \For{step $i = 1 \to \Delta \tau$}
        \State $P^{(i)} \gets \text{Model}_{\theta}(\mathbf{E}^{(i-1)})$
        \State $\mathcal{L}^{(i)} \gets \text{CrossEntropy}(P^{(i)}_{\mathcal{B}}, Y_{\mathcal{B}})$
        \State Update states $\mathbf{Z}^{(i)}$, embeddings $\mathbf{E}^{(i)}$ and decode tokens $V^{(i)}$ as in Algorithm~\ref{alg:progressive-inference}
        \State $\theta \gets \theta - \eta \nabla_\theta \mathcal{L}^{(i)}$
    \EndFor
\EndFor
\State \Return $\theta$
\end{algorithmic}
\end{algorithm*}

Algorithm~\ref{alg:continuous-sft} presents the continuous trajectory supervision procedure. The core design philosophy is to bridge the bias between the static training of traditional MDLMs and the iterative nature of EvoToken inference. 
We introduce a training-time simulation of the inference trajectory by partitioning the target block $\mathcal{B}$ into $S_{soft}$ and $S_{mask}$, and we initialize $S_{soft}$ positions with their corresponding ground-truth (GT) embeddings $e^{y_k}$. This multi-step refinement loop, repeated for $\Delta \tau$ steps within each training iteration, ensures that the model parameters $\theta$ are optimized not just for single-step recovery, but for the evolutionary path. This continuous optimization allows the model to effectively learn the progressive refinement over successive iterations.



\section{More Implementation Details}
\label{sec:more-implementation-details}

\subsection{Detailed Descriptions of Training Dataset and Evaluation Benchmarks}
\label{sec:more-implementation-details-dataset}

We utilize a combination of high-quality instruction-tuning data and diverse mathematical benchmarks.

\noindent \textbf{Training Dataset.}
\begin{itemize}
    \item \textbf{S1K}~\citep{dataset-s1k}: A high-quality dataset featuring 1,000 diverse and challenging problems, each accompanied by distilled reasoning traces and solutions to facilitate complex chain-of-thought reasoning.
\end{itemize}

\noindent \textbf{Evaluation Benchmarks.} 
We evaluate the performance of our model across the following four benchmarks, covering a spectrum of arithmetic and logical difficulty.
\begin{itemize}
    \item \textbf{Countdown}~\citep{dataset-countdown}: A combinatorial arithmetic task that requires models to reach a target value using a specific set of numbers and basic operators.
    \item \textbf{GSM8K}~\citep{dataset-gsm8k}: A collection of 8.5K grade-school math problems requiring 2--8 steps of multi-step arithmetic reasoning.
    \item \textbf{MATH500}~\citep{dataset-math500}: A subset of 500 challenging high-school competition-level problems selected from the MATH dataset.
    \item \textbf{SVAMP}~\citep{dataset-svamp}: A benchmark of 1K elementary math word problems designed to test model robustness against linguistic variations in narratives.
\end{itemize}

\subsection{Training Configurations for Continuous Trajectory Supervision}
\label{sec:more-implementation-details-continuous-sft}

To fine-tune the model under the progressive token evolution mechanism, we employ the following training configurations. The model is trained on the S1K dataset~\citep{dataset-s1k} for a total of 10k steps.

\begin{itemize}
    \item \textbf{LoRA Configuration:} The LoRA adapter is applied to the query, key, and value projections. We set the rank $r=128$, LoRA alpha $\alpha=256$, and a dropout rate of $0.05$, with no bias parameters tuned.
    \item \textbf{Optimization Settings:} We use a learning rate of $1e-5$ with a total batch size of $8$.
    \item \textbf{Sequence Handling:} The maximum sequence length is truncated at $1,024$ tokens.
    \item \textbf{Continuous Simulation:} Following our proposed framework, the number of continuous simulation steps $\Delta \tau$ is set to $4$. For blockwise processing, the current block size is fixed at $512$. During training, the number of transition tokens $|S|$ is dynamically sampled from the set $\{1, 2, 4, 8\}$ to enhance the model's robustness across varying generation densities. The mixing ratio is stochastically sampled from a uniform distribution $\mathcal{U}(0.5, 1.0)$.
\end{itemize}

\subsection{Inference and Evaluation Setup}
\label{sec:more-implementation-details-inference-eval}

To ensure a fair and reproducible evaluation, we standardize our inference parameters across all datasets. We employ a decoding temperature of $0.5$ and fix the random seed to $42$ to eliminate stochastic variance. 
For the proposed refinement mechanism, we perform a grid search for the hyperparameter $\alpha$ within the candidate set $\{0.5, 0.6, 0.7, 0.8, 0.9\}$ and report the performance associated with the optimal $\alpha$ for each setting.
For evaluations based on Dream-Instruct-7B~\citep{dream2025}, we set the default generation length to 256 with $128$ NFEs. 
For evaluations based on D2F-LLaDA~\citep{d2f}, we set the default maximum generation length to 512.

\section{More Analyses}
\label{sec:more-analyses}

\subsection{Paradigm Consistency: Why MDLMs Rapidly Adapt to EvoToken-DLM}
\label{sec:easy-adaptation-mdlm}

We begin with a brief overview of the training process for MDLMs.
Formally, we characterize the model distribution via a diffusion process consisting of a forward and a reverse process. The forward process $q(X^t | X^0)$ gradually corrupts the initial sequence $X^0$ by independently masking tokens with a time-dependent probability $t \in [0, T]$, resulting in a partially masked sequence $X^t$ where each token is replaced by a special mask symbol with probability $t$ and remains unchanged with probability $1-\frac{t}{T}$. At $t=T$, the sequence becomes fully masked. Conversely, the reverse process learns to recover the original data distribution by iteratively predicting the masked tokens in $X^t$ as $t$ transitions from $T$ to $0$.
The training objective is defined as follows:

\begin{scriptsize}
\begin{equation}
\mathcal{L}(\theta) \triangleq -\mathbb{E}_{t, X^0, X^t}\left[\frac{T}{t} \sum_{i=1}^{N} \mathbf{1}\left[x^t_i= \langle \mathrm{mask} \rangle \right] \log p_\theta\left(x^0_i \mid X^t\right)\right],
\end{equation}
\end{scriptsize}
where $N$ denotes the sequence length.

Under this training paradigm, the model gradually develops the capability to infer the potential probability distribution of each position based on its surrounding context. Notably, although the supervision is explicitly applied only to the mask tokens, the inherent generalization of the model enables it to support inference at all positions, even those not marked by a mask. 

We further substantiate this observation through empirical analysis, as shown in Table~\ref{tab:observation-generalization-for-MDLMs}. By intentionally replacing a token at a specific position with a random token in a noisy sequence, we observe the model's ability to predict the ground-truth (GT) token at that location. Our statistical findings indicate that despite the random substitution, the GT tokens remain consistently concentrated among the top-ranked candidates in the model's output distribution. This suggests that MDLMs possess a robust, context-driven predictive mechanism that transcends the specific masking patterns encountered during training.

\begin{table}[t]
    \centering
    \begin{tabular}{lcccc}
        \toprule
        $k$ & 2 & 4 & 6 & 8 \\ 
        \midrule
        rank $\le k$  & 61.2\% & 76.8\% & 91.5\% & 94.6\% \\ 
        \bottomrule
    \end{tabular}
    \caption{Probability of the ground-truth token appearing in the Top-$k$ predictions when substituting an input token in the noisy sequence with a random token. The high hit rates demonstrate the model's robust capability to infer the correct output based solely on contextual information.}
    \label{tab:observation-generalization-for-MDLMs} 
\end{table}

The inherent capability of MDLMs to continuously predict tokens based on context aligns remarkably well with the core objective of EvoToken, which aims to refine its own tokens in real-time. This alignment facilitates a seamless transition from standard MDLMs to the EvoToken-DLM framework. Leveraging this pre-trained inductive bias, the model requires only minimal supervised fine-tuning to adapt to the EvoToken paradigm. Essentially, the foundational training of MDLMs serves as a robust prior, enabling the model to internalize the evolutionary refinement process with high training efficiency.

\subsection{Adaptation Hurdles for AR Backbones: Causal Prior Mismatch}
\label{sec:hard-adaptation-ar}

In contrast to the seamless adaptation of MDLMs, backbones pre-trained under the autoregressive (AR) paradigm face significant challenges when transitioning to the EvoToken framework. This difficulty stems from a fundamental mismatch between the AR prior and the iterative refinement nature of EvoToken. 

The primary objective of AR training is to predict the next token conditioned solely on preceding context, enforced by a unidirectional causal attention mask. This intrinsic constraint prevents the model from performing bi-directional information aggregation, which is essential for iteratively updating and refining existing tokens based on full context. While some recent works attempt to bridge this gap by fine-tuning AR models into diffusion-like or blockwise models (e.g., SDAR~\cite{sdar}), these models often remain heavily tethered to their original causal priors. Consequently, adapting such models to the EvoToken paradigm necessitates substantial training resources to override the deep-seated unidirectional bias. Due to these significant computational overheads, we do not conduct further exploration on these AR-based variants.

\begin{figure}[t]
    \centering
    \includegraphics[width=\linewidth]{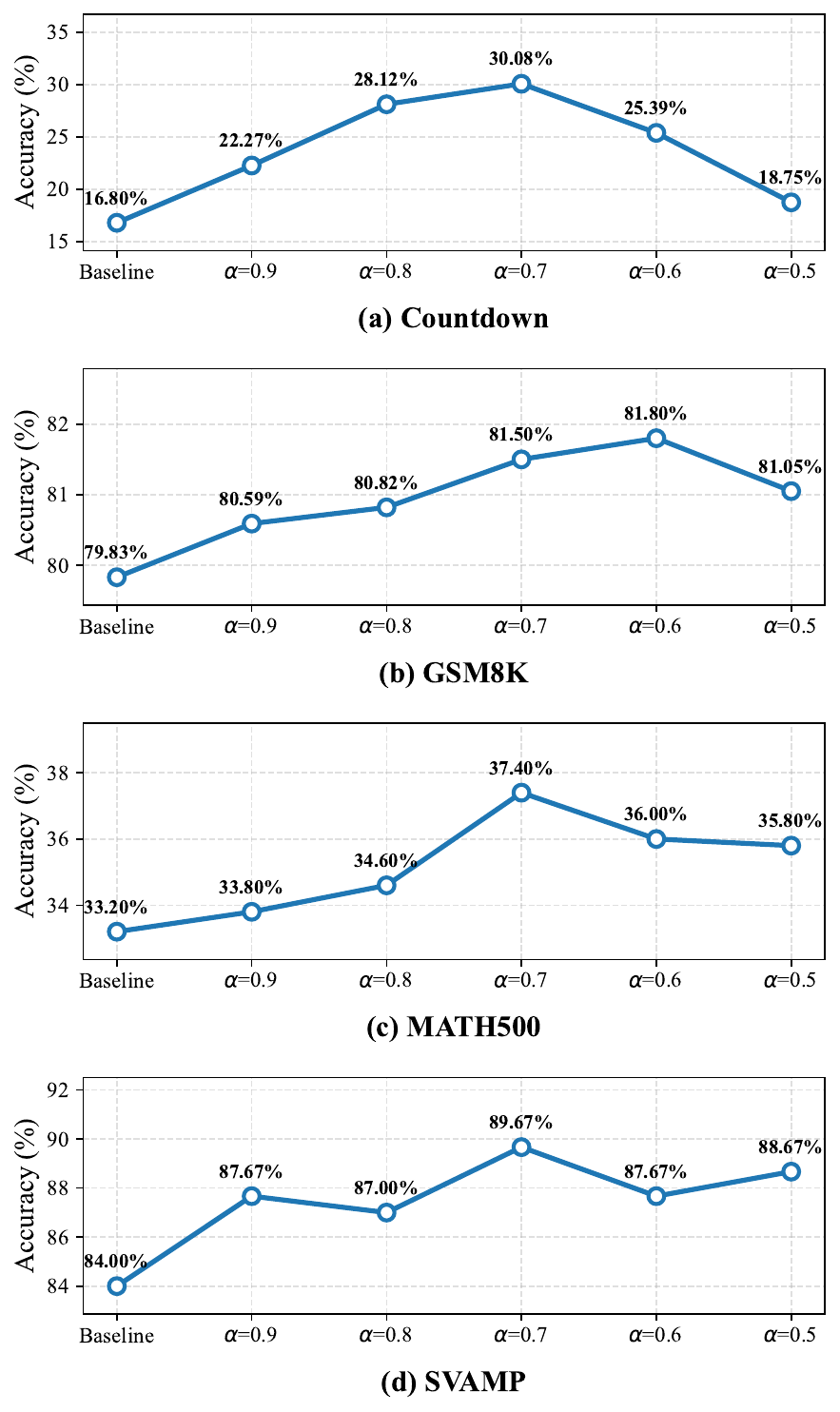}
    \vspace{-8mm}
    \caption{Performance comparison between the baseline and EvoToken-DLM with different $\alpha$ settings on various datasets.}
    \label{fig:alpha-comparison}
\end{figure}

\begin{figure*}[t]
    \centering
    \includegraphics[width=\linewidth]{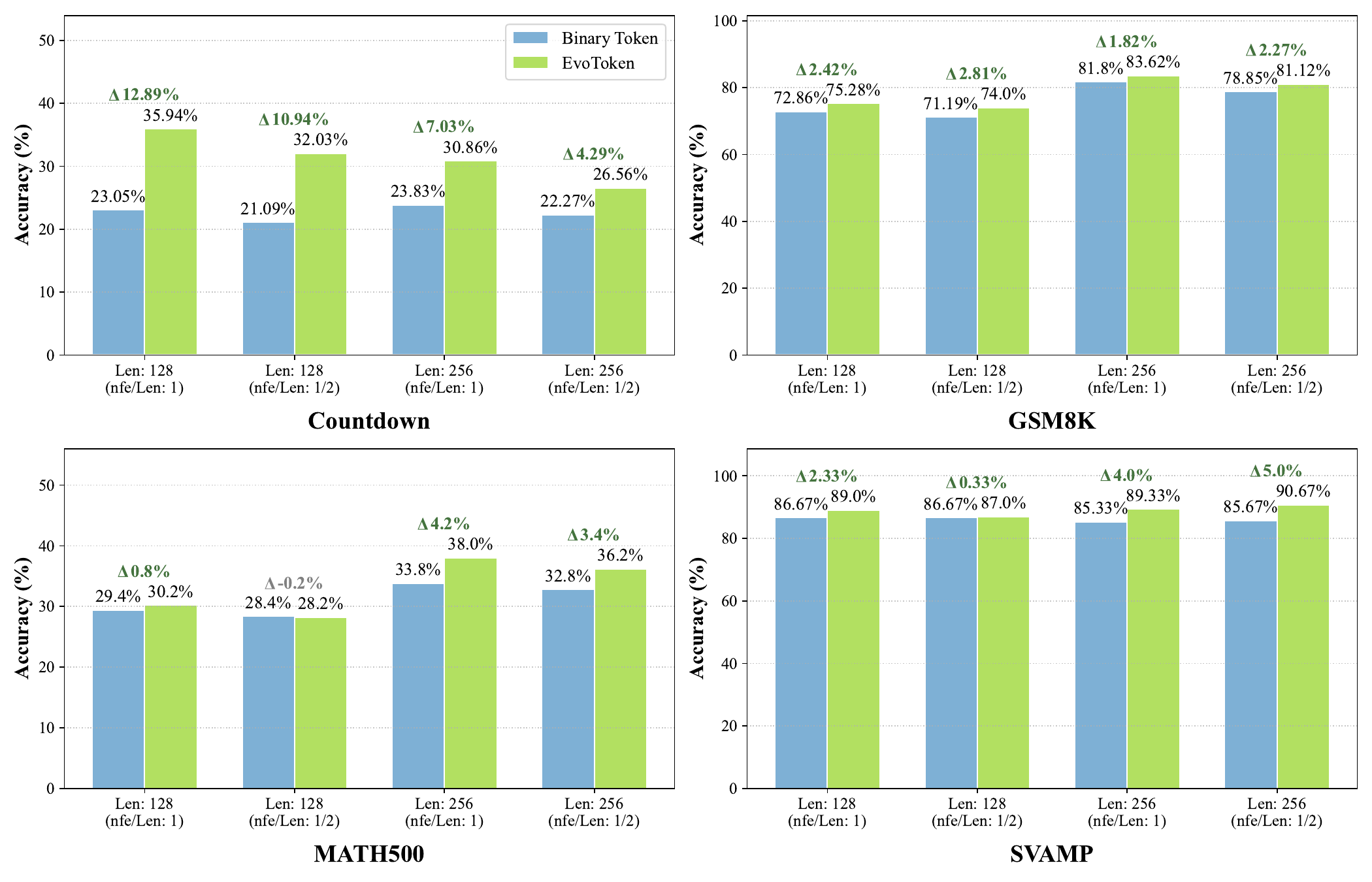}
    \vspace{-8mm}
    \caption{Additional comparison of EvoToken and the binary masking baseline based on another pretrained model LLaDA-1.5. We apply continuous trajectory supervision and evaluate performance on various benchmarks.}
    \label{fig:additional-results-1.5-all}
\end{figure*}

\section{More Experimental Results}
\label{sec:more-experimental-results}

\subsection{Additional Results Based on LLaDA-1.5}
\label{sec:additional-results-llada-1.5}

To verify the generalizability of EvoToken-DLM across different model versions, we extend our framework to the LLaDA-1.5 pretrained model via continuous trajectory supervision. Figure~\ref{fig:additional-results-1.5-all} illustrates the performance gains across multiple benchmarks. The results confirm that the improvements provided by our progressive evolution mechanism are generalizable. EvoToken-DLM consistently enhances reasoning precision compared to the binary masking baseline, even based on another pretrained model.

\subsection{Ablation Study on the Mixing Ratio \texorpdfstring{$\alpha$}{alpha}}
\label{sec:ablation-mixing-ratio}

We investigate the sensitivity of EvoToken-DLM to the mixing ratio $\alpha$ in Figure~\ref{fig:alpha-comparison}. The results demonstrate that our framework maintains remarkably stable performance across the $\alpha \in [0.5, 0.9]$ range on all four evaluation datasets, highlighting its substantial algorithmic robustness. EvoToken-DLM effectively captures the essential refinement signals even under varying fusion intensities. The fact that performance remains consistent across diverse benchmarks, ranging from simple arithmetic to complex reasoning, underscores the robustness of the mechanism.

\subsection{Additional Results with Different Block Sizes}
\label{sec:additional-results-different-block-size}

\begin{table*}[t] 
\centering
\caption{Performance comparison on the Countdown, GSM8K, MATH500 and SVAMP datasets across various block sizes based on LLaDA-Instruct-8B. EvoToken-DLM is initialized from LLaDA-Instruct-8B and fine-tuned for 10k steps using continuous trajectory supervision. Comparisons are conducted against both the baseline model and the sft-baseline. All experiments are conducted with a generation length of 256 and an NFE of 128.}
\label{tab:various-blocksizes} 
\vspace{-4mm}
\setlength{\tabcolsep}{10pt} 
\scalebox{0.8}{ 
\begin{tabular}{ll|cccc}
    \toprule
    \textbf{Block Size} & \textbf{Method} & \textbf{Countdown} & \textbf{GSM8K} & \textbf{MATH500} & \textbf{SVAMP} \\
    \midrule
    \multirow{4}{*}{\textbf{8}} 
    & Baseline & 12.11\% & 74.30\% & 32.00\% & 85.00\% \\
    \cmidrule(lr){2-6}
    & FT-Base & 17.97\% & 76.50\% & 31.60\% & 87.33\% \\
    \cmidrule(lr){2-6}
    & \multirow{2}{*}{\textbf{EvoToken}} & 22.66\% & 79.00\% & 34.60\% & 88.00\% \\
    &  & \textcolor{green!70!black}{+10.55\%} & \textcolor{green!70!black}{+4.70\%} & \textcolor{green!70!black}{+2.60\%} & \textcolor{green!70!black}{+3.00\%} \\

    \midrule
    \midrule
    \multirow{4}{*}{\textbf{16}} 
    & Baseline & 15.23\% & 75.21\% & 32.60\% & 84.33\% \\
    \cmidrule(lr){2-6}
    & FT-Base & 16.80\% & 79.08\% & 32.00\% & 88.67\% \\
    \cmidrule(lr){2-6}
    & \multirow{2}{*}{\textbf{EvoToken}} & 26.95\% & 80.59\% & 37.20\% & 90.00\% \\
    &  & \textcolor{green!70!black}{+11.72\%} & \textcolor{green!70!black}{+5.38\%} & \textcolor{green!70!black}{+4.60\%} & \textcolor{green!70!black}{+5.67\%} \\

    \midrule
    \midrule
    \multirow{4}{*}{\textbf{32}} 
    & Baseline & 16.41\% & 77.63\% & 32.20\% & 86.67\% \\
    \cmidrule(lr){2-6}
    & FT-Base & 16.80\% & 78.62\% & 31.20\% & 87.00\% \\
    \cmidrule(lr){2-6}
    & \multirow{2}{*}{\textbf{EvoToken}} & 30.08\% & 82.03\% & 36.40\% & 92.33\% \\
    &  & \textcolor{green!70!black}{+13.67\%} & \textcolor{green!70!black}{+4.40\%} & \textcolor{green!70!black}{+4.20\%} & \textcolor{green!70!black}{+5.66\%} \\
    
    \midrule
    \midrule
    \multirow{4}{*}{\textbf{64}} 
    & Baseline & 19.92\% & 76.42\% & 33.80\% & 86.67\% \\
    \cmidrule(lr){2-6}
    & FT-Base & 19.14\% & 78.01\% & 34.60\% & 87.33\% \\
    \cmidrule(lr){2-6}
    & \multirow{2}{*}{\textbf{EvoToken}} & 31.64\% & 83.40\% & 38.00\% & 91.00\% \\
    &  & \textcolor{green!70!black}{+11.72\%} & \textcolor{green!70!black}{+6.98\%} & \textcolor{green!70!black}{+4.20\%} & \textcolor{green!70!black}{+4.33\%} \\
    \bottomrule
\end{tabular}
} 
\end{table*}

To further evaluate the robustness and scalability of EvoToken-DLM, we conduct experiments across a range of block sizes $\in \{8, 16, 32, 64\}$. As shown in Table~\ref{tab:various-blocksizes}, our method consistently outperforms both the original LLaDA-Instruct-8B baseline and the FT-baseline across all tested configurations and datasets.

The experimental results highlight several key observations:
\begin{itemize}
    \item \textbf{Universal Performance Gains:} EvoToken-DLM achieves significant accuracy improvements regardless of the block size. For instance, in the Countdown task, we observe absolute gains ranging from $+10.55\%$ to $+13.67\%$ over the baseline, demonstrating that our approach effectively enhances the model's reasoning capabilities across various inference granularities.
    \item \textbf{Robustness to Block Size Variations:} EvoToken-DLM demonstrates high resilience to changes in block size, yielding stable and superior results across all tested discretization settings. This suggests that the learned token evolution patterns are agnostic to specific block partitions.
    \item \textbf{Consistency across Tasks:} The superiority of our method is consistently maintained across diverse benchmarks, from symbolic reasoning (Countdown) to complex mathematical problem-solving (GSM8K, MATH500), further validating the generalizability of our approach in various downstream scenarios.
\end{itemize}

Overall, these results underscore that EvoToken-DLM is not finely tuned for a specific inference setting but rather provides a fundamental enhancement to the underlying diffusion generation process.

\begin{figure*}[t]
    \centering
    \includegraphics[width=\linewidth]{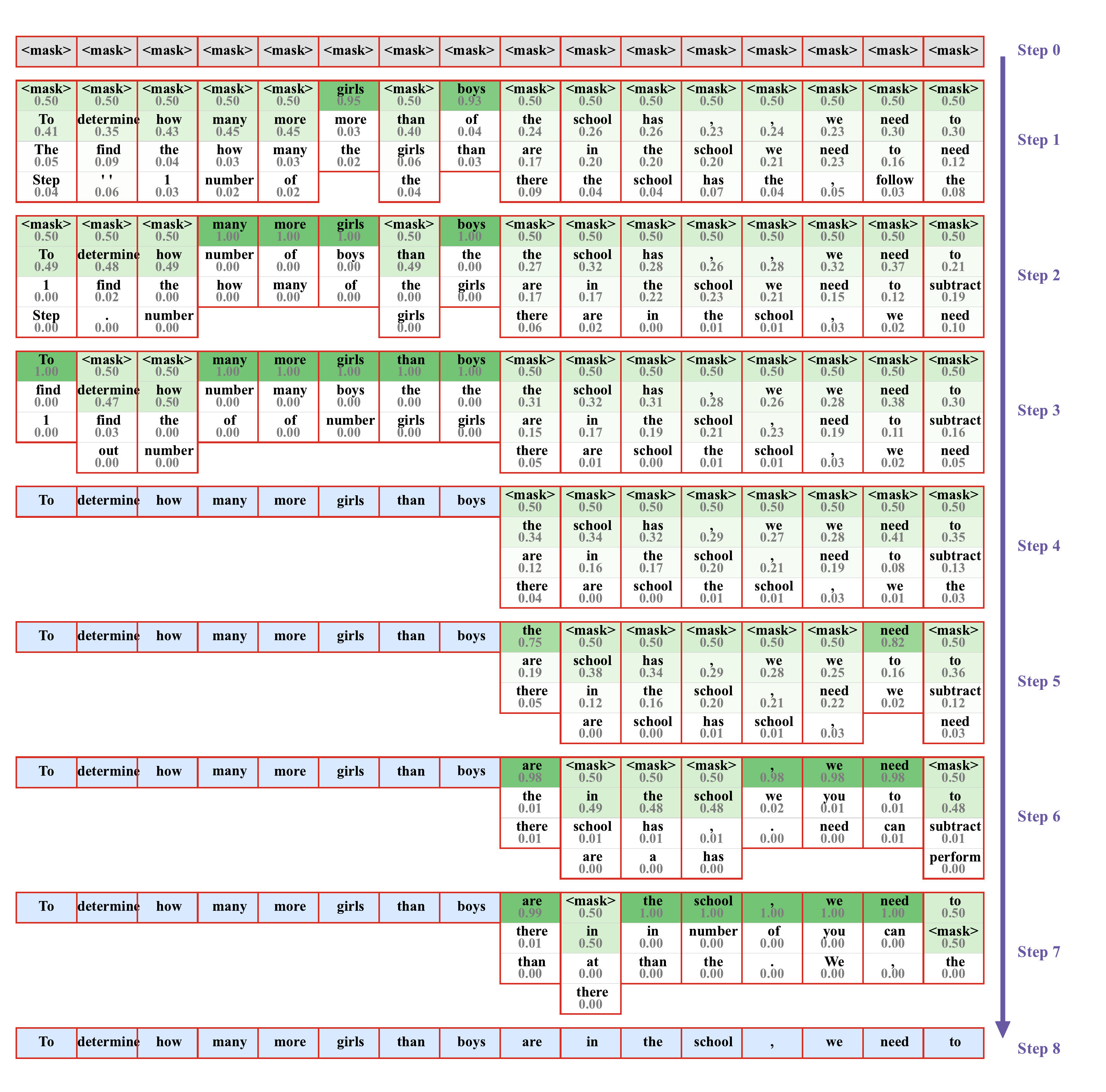}
    \vspace{-8mm}
    \caption{
    Additional inference example with a block size of 8, showing intermediate refinement states for a selected subsequence across successive steps. We showcase the refinement states for the first 16 output tokens based on the prompt: "In a school there are 569 girls and 236 boys.How many more girls than boys does the school have?".
    }
    \label{fig:refine-example-additional-2}
\end{figure*}

\begin{figure*}[t] 

    \subsection{Qualitative Comparisons for EvoToken-DLM and MDLMs}
    \label{sec:qualitative-comparisons}
    
    \begin{examplebox}{Example 1}
        \textbf{Question:} \textit{Mattison is 60 inches tall. His two-story house is 7 times as high as he is. The floor to the ceiling on the first floor is thrice as high as he is. How high is the roof from the second floor's floor in inches?}
        
        \textbf{GT:} 240.0
        \vspace{2mm}
        \begin{description}
            \item[\textsc{MDLMs:}] 

                To determine the height of the roof from the second floor's floor, we need to follow these steps:
                
                1. Calculate the height height of Mattison's two-story house.
                2. Calculate the height of the floor to the ceiling on the first floor.
                3. Determine the height of the second floor from the ground.
                4. Calculate the height of the roof from the second floor's floor.
                
                1. The height of Mattison's two-story house is 7 times his height.
                   \[
                   \text{Height of the house} = 7 \times 60 = 420 \text{ inches}
                   \]
                
                2. The height of the floor to the ceiling on the first floor is 3 times his height.
                   \[
                   \text{Height of floor to ceiling} = 3 \times 60 = 180 \text{ inches}
                   \]
                
                3. The height of the second floor from the ground is the height of the first floor minus the height of the first floor.
                   \[
                   \text{Height of second floor} = 420 - 180 = 240 \text{ inches}
                   \]
                
                4. The height of the roof from the second floor's floor is the height of the second floor minus the height of the first floor.
                   \[
                   \text{Height of roof} = 240 - 180 = 60 \text{ inches}
                   \]
                
                \detokenize{<answer>}
                \boxed{60}
                \detokenize{</answer>}

            \item[\textsc{EvoToken-DLM (Ours):}] 
                Solution:
                
                Let H1 be the height of the first floor.
                
                Let H2 be the height of the second floor.
                
                Let H be the the height of the house.
                
                We are given:
                Mattison's height, 60 inches.
                The height of the house is 7 times Mattison's height, so H = 7 * 60 = 420 inches.
                The floor to the ceiling on the first floor is 3 times Mattison's height, so H1 = 3 * 60 = 180 inches.
                
                The height of the house is the sum of the height of the first floor and the height of the second floor:
                
                H = H1 + H2
                
                420 = 180 + H2
                
                H2 = 420 - 180
                
                H2 = 240 inches
                
                The height of the roof from the second floor's floor is the height of the second floor, which is 240 inches.
                
                Final Answer: The final answer is $\boxed{240}$\detokenize{</reasoning>}

                \detokenize{<answer>}\boxed{240}\detokenize{</answer>}
                
        \end{description}
    \end{examplebox}

\end{figure*}

\begin{figure*}[t] 
    
    \begin{examplebox}{Example 2}
        \textbf{Question:} \textit{A class of 30 students recently took a test.  If 20 students scored 80, 8 students scored 90, and 2 students scored 100, then what was the class average (the mean) on this test?}
        
        \textbf{GT:} 84
        \vspace{2mm}
        \begin{description}
            \item[\textsc{MDLMs:}] 
                To find the class average (the mean) on the test, we need to calculate the total score of all students and then divide by the total number of students.
    
                1. Calculate the total score for the students who scored students scored 80:
                   \[
                   20 \times 80 = 1600
                   \]
                
                2. Calculate the total score for the students who scored 90:
                   \[
                   8 \times 90 = 720
                   \]
                
                3. Calculate the total score for the students who scored 100:
                   \[
                   2 \times 100 = 100
                   \]
                
                4. Add these totals together to get the total score:
                   \[
                   1600 + 720 + 100 = 2420
                   \]
                
                5. Divide the total score by the total number of students to find the mean:
                   \[
                   \frac{2420}{30} = 80.67
                   \]
                
                \detokenize{</, reasoning>}
                
                \detokenize{<answer>}
                \boxed{80.67}
                \detokenize{</answer><|eot_id|>}

            \item[\textsc{EvoToken-DLM (Ours):}] 
                Solution:

                To find the average score, we need the total sum of all scores and then divide by the number of students.
                
                20 students scoring 80 : 20 * 80 = 1600
                
                8 students scoring 90 : 8 * 90 = 720
                
                2 students scoring 100 : 2 * 100 = 200
                
                Total sum of scores = 1600 + 720 + 200 = 2520
                
                Number of students = 30
                
                Average = Total sum of scores / Number of students = 2520 / 30 = 84
                
                Final Answer: The final answer is $\boxed{84}$\detokenize{</reasoning>}
                
                \detokenize{<answer>}\boxed{84}\detokenize{</answer>}

        \end{description}
    \end{examplebox}

\end{figure*}

\begin{figure*}[t] 
    
    \begin{examplebox}{Example 3}
        \textbf{Question:} \textit{Using only the provided numbers, create an arithmetic expression that evaluates to exactly the provided target number. You may use the operations +, -, *, and / as needed, but each number must be used exactly once. Numbers: [92, 43, 58]; Target: 77.}
        
        \textbf{GT:} 92 - 58 + 43
        \vspace{2mm}
        \begin{description}
            \item[\textsc{MDLMs:}] 
                To create an arithmetic expression that evaluates to 77 using the numbers 92, 43, and 58, we need to find a combination of operations (+, -, *, /) that will result in 77. Let's start by considering the different operations:

                1. **Addition and Subtraction**:
                   - \(92 + 43 = 135\)
                   - \(92 - 43 = 49\)
                   - \(92 + 58 = 150\)
                   - \(92 - 58 = 34\)
                
                2. **Multiplication**:
                   - \(92 \times 43 = 3956\)
                   - \(92 \times 58 = 5376\)
                   - \(43 \times 58 = 2454\)
                
                3. **Division**:
                   - \(92 \div 43 \approx 2.097\)
                   - \(92 \div 58 \approx 1.587\)
                   - \(43 \div 58 \approx 0.732\)
                
                We need to find a combination of these operations that results in 77. Let's try different combinations:
                
                1. **Multiplication and**:  
                   \(996 - 43 = 953\)  
                   \(953 - 58 = 895\)  
                   \(895 - 43 = 852\)  
                   \(852 - 58 = 794\)  
                   \(794 - 43 = 751\)  
                   \(751 - 58 = 693\)  
                   \(693 - 43 = 650\)  
                   \(650 - 58 = 592\)  
                   \(592 - 43 = 549\)  
                   \(549 - 58 = 77\)  
                
                Thus, the expression \(592 - 43 - 58 = 77\)
                
                \detokenize{</answer>}
                \boxed{592 - 43 - 58}
                \detokenize{</answer>}

            \item[\textsc{EvoToken-DLM (Ours):}] 
                We need to find an arithmetic expression using the numbers 92, 43, and 58 that evaluates to 77. We operations allowed are +, -, *, and / Each number must be used exactly once.

                Let's try different combinations of operations:
                
                1. **Subtraction**: 92 - 43 = 49, 49 - 58 = -9, -9 + 77 = 68 (not target)
                2. **Multiplication**: 92 * 43 = 3916, 3916 - 58 = 3858, 3858 - 77 = 3981 (not target)
                3. **Division**: 92 / 43 = 2, 2 * 58 = 116, 116 - 77 = 39 (not target)
                
                Let's try a different approach:
                
                 the. 92 - 43 = 49, 49 - 58 = -9 (not target)
                
                 the. 92 / 43 = 2, 92 - 58 = 34 (not target)
                
                 the. 92 - 58 = 34, 34 + 43 = 77 (target)
                
                The correct expression is 92 - 58 + 43.
                \detokenize{</reasoning>}
                
                \detokenize{<answer>}\boxed{92 - 58 + 43}\detokenize{</answer>}

        \end{description}
    \end{examplebox}

\end{figure*}

\begin{figure*}[t] 
    
    \begin{examplebox}{Example 4}
        \textbf{Question:} \textit{Faye was placing her pencils into rows with 22 pencils in each row. She had 6 packs of pencils each one having 14 pencils.How many pencils does she have?}
        
        \textbf{GT:} 84.0
        \vspace{2mm}
        \begin{description}
            \item[\textsc{MDLMs:}] 
                To determine the total number of pencils Faye has, we need to add the number of pencils in the rows to the number of pencils in the packs. First, we calculate the total number of pencils in the rows by Since there are 22 pencils in each row and Faye has multiple rows, we don't need to exact number of rows for this calculation. Next, we calculate the total number of pencils in the packs by multiplying the number of packs by the number of pencils in each pack. Faye has 6 packs, and each pack contains 14 pencils, so the total number of pencils in the packs is \(6 \times 14 = 84\). Finally, we add the number of pencils in the rows to the number of pencils in the packs to get the total number of pencils.
                \detokenize{</reasoning>}
                
                \detokenize{<answer>}
                \boxed{116}
                \detokenize{</answer>}

            \item[\textsc{EvoToken-DLM (Ours):}] 
                We need to determine the total number of pencils Faye has. We know that she has 6 packs with 14 pencils each.

                 number of pencils = number of packs * number of pencils per pack
                 number of pencils = 6 * 14 = 84
                
                 Faye is placing her pencils into rows with 22 pencils in each row. We need to find out how many rows she can make with 84 pencils.
                
                 number of rows = number of pencils / number of pencils per row
                 number of rows = 84 / 22 = 4
                
                However the question asks for the total number of pencils, which is 84.
                
                Final Answer: The final answer is $\boxed{84}$
                \detokenize{</reasoning>}
                
                \detokenize{<answer>}\boxed{84}\detokenize{</answer>}
                
        \end{description}
    \end{examplebox}

\end{figure*}

\end{document}